\pdfoutput=1

\documentclass[11pt]{article}

\usepackage[preprint]{acl}

\usepackage{times}
\usepackage{latexsym}

\usepackage[T1]{fontenc}

\usepackage[utf8]{inputenc}

\usepackage{microtype}

\usepackage{inconsolata}

\usepackage{graphicx}
\usepackage{amsmath}
\usepackage{times}
\usepackage{framed}
\usepackage{ragged2e}
\usepackage{bm}
\usepackage{soul}
\usepackage{latexsym}
\usepackage{subfigure}
\usepackage{amsthm}
\usepackage{graphicx}
\usepackage{float}
\usepackage{longtable}
\usepackage{booktabs} 
\usepackage{multirow}
\usepackage{multicol}
\usepackage{amssymb}
\usepackage{dashbox}%
\usepackage{multirow}
\usepackage{textcomp}
\usepackage{tcolorbox}
\usepackage{bbding}
\usepackage{bbm}
\usepackage[ruled, vlined, linesnumbered]{algorithm2e}
\usepackage{mathtools}
\usepackage{adjustbox}
\usepackage[ruled,vlined]{algorithm2e}
\usepackage{color, soul}
\usepackage{pifont}
\usepackage{array}
\newcolumntype{P}[1]{>{\centering\arraybackslash}p{#1}}
\newcolumntype{M}[1]{>{\centering\arraybackslash}m{#1}}
\usepackage{cleveref}
\crefname{section}{§}{§§}
\Crefname{section}{§}{§§}
\crefname{figure}{Figure}{Figure}
\Crefname{figure}{Figure}{Figure}
\crefname{table}{Table}{Table}
\Crefname{table}{Table}{Table}
\usepackage{tabularx}
\usepackage{float} 
\newcommand\ourmodel{\textsc{Crab}\xspace}
\newcommand\ourdata{\textsc{Crab}\xspace}
\newcommand\llamacrab{Llama3\textsubscript{\textsc{Crab}}\xspace}
\newcommand\mistralcrab{Mistral\textsubscript{\textsc{Crab}}\xspace}

\usepackage{colortbl}

\title{Constraint Back-translation Improves Complex Instruction Following \\of Large Language Models}

\author{Yunjia Qi\thanks{Equal contribution.}, Hao Peng \hspace{-3pt}$^*$, Xiaozhi Wang, Bin Xu, Lei Hou, Juanzi Li \\
        Department of Computer Science and Technology, BNRist, Tsinghua University \\ \texttt{\{qyj23, peng-h24\}@mails.tsinghua.edu.cn}}

\begin{document}
\maketitle
\begin{abstract}



Large language models (LLMs) struggle to follow instructions with complex constraints in format, length, etc. Following the conventional instruction-tuning practice, previous works conduct post-training on complex instruction-response pairs generated by feeding complex instructions to advanced LLMs. However, even advanced LLMs cannot follow complex instructions well, thus limiting the quality of generated data. In this work, we find that \textit{existing datasets inherently contain implicit complex constraints} and propose a novel data generation technique, \textit{constraint back-translation}. Specifically, we take the high-quality instruction-response pairs in existing datasets and only adopt advanced LLMs to add complex constraints already met by the responses to the instructions, which naturally reduces costs and data noise. In the experiments, we adopt Llama3-70B-Instruct to back-translate constraints and create a high-quality complex instruction-response dataset, named \ourmodel. We present that post-training on \ourmodel improves multiple backbone LLMs' complex instruction-following ability, evaluated on extensive instruction-following benchmarks. We further find that constraint back-translation also serves as a useful auxiliary training objective in post-training. Our code, data, and models will be released to facilitate future research\footnote{\url{https://github.com/THU-KEG/Crab}}.


\end{abstract}


\section{Introduction}

\begin{figure}[t]
    \centering
    \includegraphics[width=1.0\linewidth]{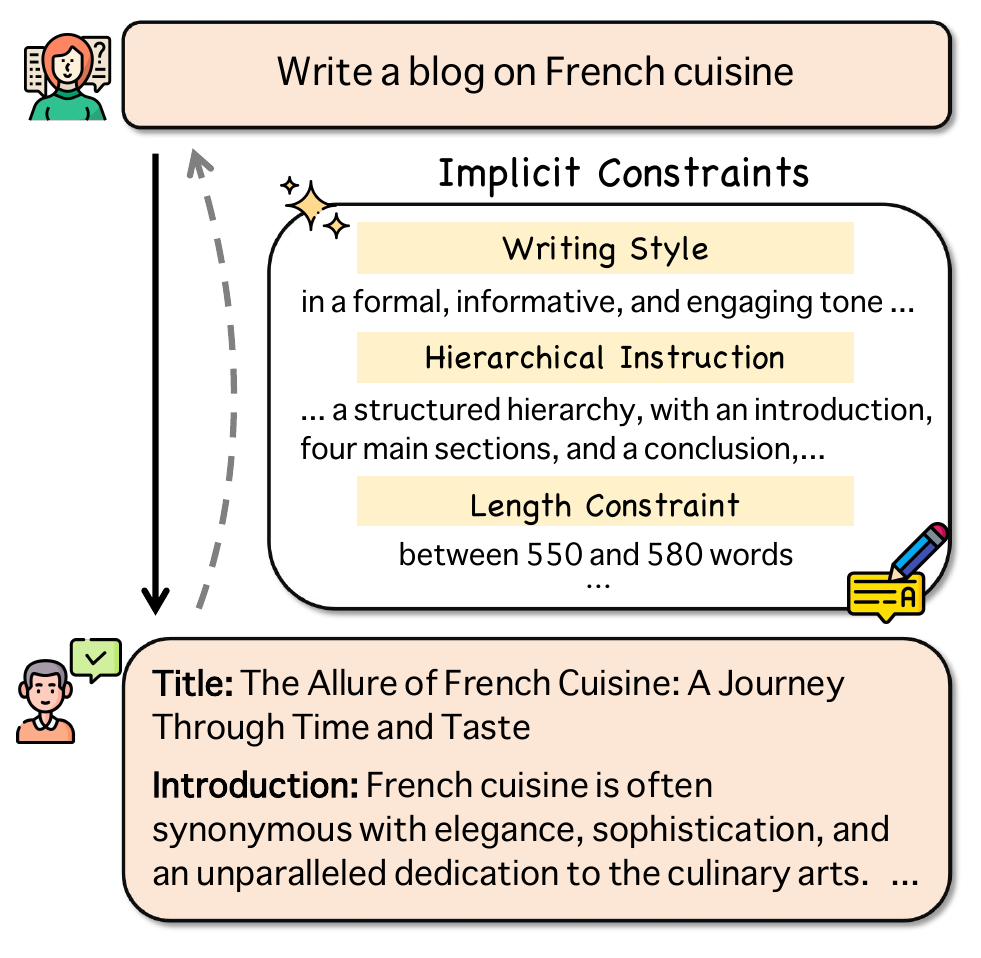} 
    \caption{Existing datasets inherently include implicit satisfied complex constraints in the responses.
    }
    \label{fig:fig1}
\end{figure}



Large language models (LLMs) have achieved remarkable performance in numerous natural language processing tasks~\citep{zhao2023survey,gpt4,yang2024qwen2,dubey2024llama,team2024gemma}. 
However, they still fall short in following instructions with complex constraints~\citep{zhou2023instruction, jiang-etal-2024-followbench, qin2024infobench}, such as length constraints shown in Figure~\ref{fig:fig1}, which limits their effectiveness and usability.


To enhance the instruction-following ability of LLMs, the standard practice is to post-train the targeted LLM on a large set of instruction-response data pairs. For the complex instruction-following with multiple constraints, existing efforts~\citep{sun2024conifer,he2024complex} synthesize complex datasets by adding multiple constraints to existing instructions and generating responses with advanced LLMs like GPT-4~\citep{gpt4}. While this data generation pipeline is straightforward and widely adopted, even the most capable LLMs cannot follow complex instructions well~\citep{jiang-etal-2024-followbench, qin2024infobench}, which limits the quality of generated data and necessites laborious filtering. The status quo urges the development of automatic data generation methods relying less on existing LLMs' complex instruction-following abilities.

Our key observation is that \textit{existing datasets inherently include implicit complex constraints} so that we can reuse the widely-available high-quality instruction-following datasets~\citep{xu2023wizardlm,taori2023alpaca,mukherjee2023orca,kopf2024openassistant} to synthesize complex instruction-response pairs. As shown in Figure~\ref{fig:fig1}, although the original concise instruction does not explicitly specify constraints like writing style or length, the response already satisfies some constraints in multiple dimensions. Therefore, we can efficiently create high-quality complex instruction-response pairs from existing datasets by generating constraints from responses and adding them to instructions.
We dub this data generation method as \textit{constraint back-translation}. It only requires discovering the constraints already met by responses rather than following the complex instructions with multiple constraints, which significantly reduces requirements for model capability. As a result, it is both cost-effective and capable of producing high-quality data with limited noise. We also find that constraint back-translation can serve as a useful auxiliary training objective in post-training, dubbed as the \textit{reverse training} technique. Specifically, we use instructions and responses as inputs to train the model to output constraints in post-training. The intuition is that reverse training may enhance the model's understanding of constraints and improve its efficacy~\citep{golovneva2024reverse}.

We adopt Llama3-70B-Instruct~\citep{dubey2024llama} to back-translate constraints from a collection of existing data, generating a large-scale complex instruction-following dataset, named \ourmodel. Specifically, we sample a total of $13,500$ instances from existing high-quality instruction-following datasets~\citep{alpaca-gpt4,Orca-Chat,xu2023wizardlm,kopf2024openassistant} as the seed data, and manually define a scope of common constraints. We then use the original instruction, response, and constraint scope as inputs to Llama3-70B-Instruct to generate the corresponding implicitly satisfied constraints. Following previous works~\citep{sun2024conifer, he2024complex}, we train the LLMs using the mixture of \ourmodel and ShareGPT dataset~\citep{chiang2023vicuna}, and we jointly adopt standard supervised fine-tuning and reverse-training on \ourmodel. In the experiments, we select the capable open-source LLMs Llama3 8B~\citep{dubey2024llama} and Mistral 7B~\citep{jiang2023mistral} as backbone models and evaluate the complex instruction-following abilities of our models against various baselines on IFEval~\citep{zhou2023instruction} and FollowBench~\citep{jiang-etal-2024-followbench}. The results demonstrate that training on \ourmodel significantly enhances LLM performance in complex instruction following. We also conduct evaluation for general instruction-following abilities on AlpacaEval~\citep{alpaca_eval} and find that our models achieve even larger improvements to previous works focusing on enhancing complex instruction-following abilities like Conifer~\citep{sun2024conifer}. This indicates that constraint back-translation yields higher general data quality than previous techniques relying on the ability of advanced LLMs. Ablation studies further validate the efficacy of our \ourmodel dataset and reverse training approach. Finally, we discuss the advantages and challenging scenarios for our constraint back-translation method with experiments

In summary, our contributions are threefold: 
(1) We propose constraint back-translation, a cost-effective and high-quality data construction method for complex instruction following.
(2) We construct \ourmodel, a high-quality complex instruction-following dataset, and design a reverse training method for developing \llamacrab, \mistralcrab models with better complex instruction-following abilities.
(3) We conduct extensive experiments to demonstrate the efficacy of \ourmodel and discuss key design choices and potential improvement opportunities to inspire future research on following complex instructions with multiple constraints.

\section{Method}

\begin{figure*}[t]
    \includegraphics[width=0.99\linewidth]{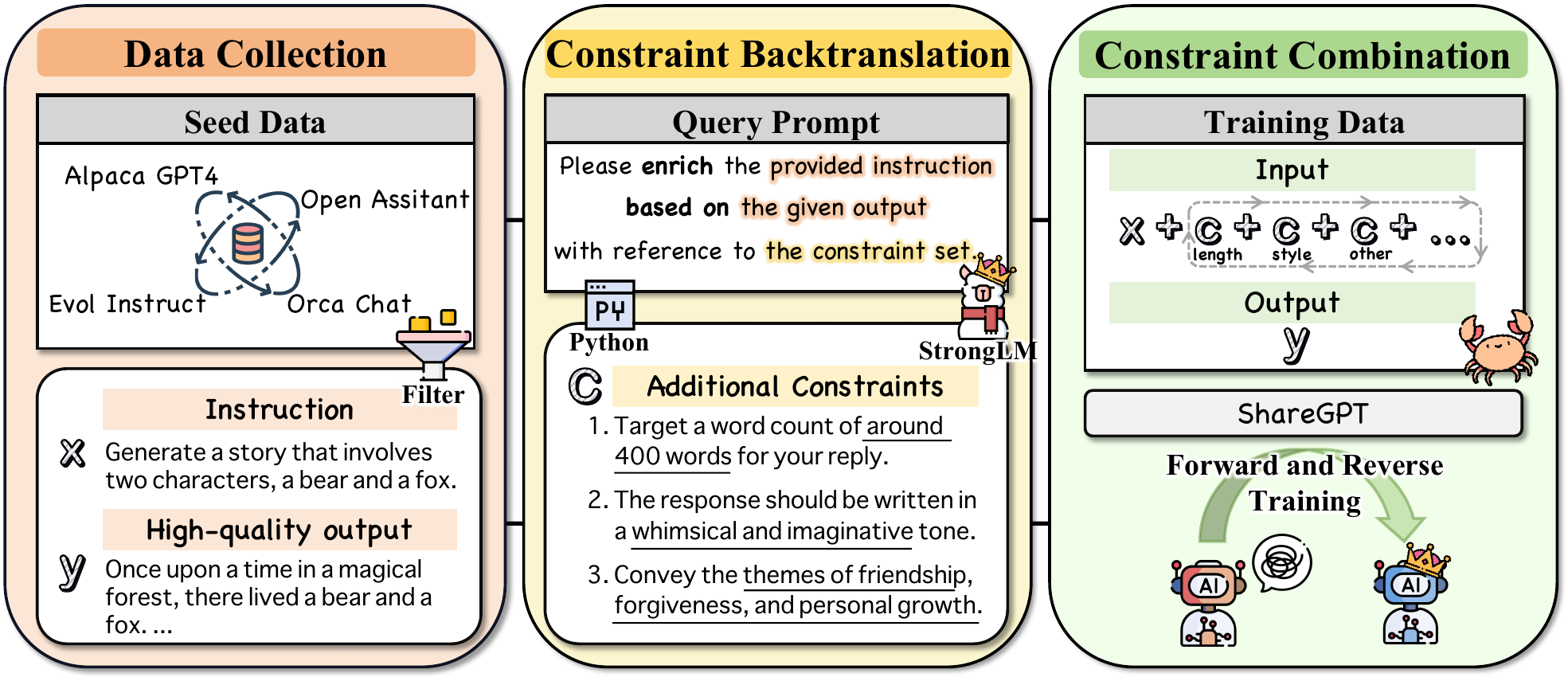} 
    \caption{The framework of constructing the proposed alignment training dataset.
    }
    \label{fig:pipeline}
\end{figure*}

This section introduces the construction process of \ourmodel (\cref{sec:data_collect}) and the training method (\cref{sec:model_training}). 


\subsection{Constructing \ourmodel}
\label{sec:data_collect}
We begin by introducing the notions. Given an instruction $x$, which typically defines a specific task, such as ``\textit{Write a blog on French cuisine}'', a set of constraints $c$, which specify conditions for the response, such as length restrictions, and a response $y$ that satisfies both the constraints $c$ and the instruction $x$, our goal is to construct a high-quality dataset of
($x, c, y$) triples. We first collect a set of high-quality ($x, y$) pairs from existing datasets and then apply constraint back-translation to generate the constraints 
$c$ for each ($x, y$) pair. The data construction process is illustrated in Figure~\ref{fig:pipeline}, which consists of three steps: data collection, constraint back-translation, and constraint combination.
In the data collection process, we collect a comprehensive set of high-quality ($x, y$) pairs from existing datasets. We then back-translate the corresponding $c$ for each ($x, y$) using Llama3-70B-Instruct and Python scripts automatically. Finally, we perform filtering and a combination of the constraints $c$ to construct \ourmodel. More details of the data construction process are shown in appendix~\ref{sec:app_data_collection}.


\paragraph{Data Collection}
%
We first collect a comprehensive set of ($x, y$) as seed data from four existing widely-used high-quality supervised fine-tuning datasets, including Alpaca GPT4~\citep{alpaca-gpt4}, Orca Chat~\citep{Orca-Chat}, Evol Instruct~\citep{xu2023wizardlm}, and OpenAssitant~\citep{kopf2024openassistant}. For Alpaca GPT4 and OpenAssistant, which contain the human annotated quality score for each instance, we use the instances with the highest quality.
Moreover, to ensure that the responses include diverse constraints implicitly, we only consider instances where the response exceeds $300$ words. 
We randomly sample the qualified instances using an examples-proportional mixture~\citep{wei2021finetuned}, resulting in a total of $4,500$ raw instances. 

\paragraph{Constraint Back-translation}
We adopt the Llama3-70B-Instruct LLM and Python scripts to back-translate constraints for 
the seed data.
We ultimately use Llama3-70B-Instruct to automatically generate constraints implicitly satisfied by the response from instruction-response pairs. To enhance the diversity of generated constraints, we manually collect $13$ commonly used constraint types{\footnote{We sample $200$ instruction-response pairs from ShareGPT to observe real-world constraint needs and summarize them.}} as examples in the prompt for constraint generation, which results in over \textbf{100} constraint types. We then use Llama3-70B-Instruct to re-verify whether the response satisfies the generated constraints and exclude the constraints that are not met.
Considering some constraints, e.g., length constraint, cannot be effectively followed by LLMs~\citep{sun2024conifer}, leading to noisy back-translation, and some constraints can be easily generated using Python scripts, we choose to adopt Python scripts for $6$ types of constraints. 
Specifically, we write and paraphrase several templates for each of these constraints. For example, for a length constraint, one template is ``Please generate a response with fewer than \texttt{<placeholder>} words but more than \texttt{<placeholder>} words''. We then use Python scripts to automatically identify the value for this constraint in the response and fill the templates to construct a constraint. For the length constraint, we randomly sample a range that includes the value to fill the template. For keyword and punctuation constraints, we randomly select corresponding items present in the response to fill the templates of constraints. 
We adopt ROUGE-L~\citep{lin2004rouge} with a threshold of $0.6$ to exclude similar constraints.
Finally, we sample and manually review $50$ instruction-response pairs and their generated constraints, finding minimal noise and high compliance between constraints and response. 

\paragraph{Constraint Combination}

Finally, we combine individual constraints to form the final constraint $c$ for each instruction. Previous studies have shown that increasing the number of constraints in the training data leads to better model performance~\citep{he2024complex}. Therefore, we enhance each instruction with a combination of multiple constraints. Specifically, we randomly sample $6$ to $8$ constraints from each instruction's constraint set generated in the previous step, shuffle their order, and recombine them into the final constraint $c$. Similar to previous work~\citep{sun2024conifer,qi2024adelie}, we add $1$ to $3$ in-context demonstrations for $50\%$ of the data. Finally, we construct \ourmodel with $13,500$ instances, containing an average of $7.1$ constraints.

\subsection{Model Training}
\label{sec:model_training}
To further enhance LLMs' understanding of complex constraints, we propose a \textit{reverse training} method that takes the instruction-response pair ($x, y$) as input to teach LLMs to generate the constraints $c$. The intuition is that correctly generating constraints requires sufficient comprehension first. Formally, the reverse training objective is to minimize $\mathcal{L}_{\text{r}}$, where $\mathcal{L}_{\text{r}} = -\log P_{\theta}(c|x, y)$ and the LLM is parameterized by $\theta$. We also adopt the standard supervised fine-tuning (SFT; \citealp{ouyang2022training}), named \textit{forward training}, to minimize $\mathcal{L}_{\text{f}}$, where $\mathcal{L}_{\text{f}} = -\log P_{\theta}(y|x, c)$. The final training objective is a combination of $\mathcal{L}_{\text{f}}$ and $\mathcal{L}_{\text{r}}$: $\mathcal{L} = 
\alpha \mathcal{L}_{\text{f}} + (1-\alpha) \mathcal{L}_{\text{r}}$.
We set $\alpha$ to $0$ during $70\%$ of the training process, and to $1$ for the remaining time. We train the LLM using a mixture of ShareGPT~\citep{chiang2023vicuna} and \ourmodel. We adopt $\mathcal{L}_{\text{f}}$ when training on ShareGPT~\citep{chiang2023vicuna}  and adopt $\mathcal{L}$ on \ourmodel.
We train a base LLM on this data and obtain the SFT version of our model.
Based on the SFT trained model, we continue training using the Direct Preference Optimization objective (DPO; \citealp{dpo}). Specifically, same as the DPO phrase by~\citet{sun2024conifer}, we use the high-quality DPO dataset UltraFeedback~\citep{cui2023ultrafeedback} to conduct further training and obtain the DPO version of our model. More training details are presented in appendix~\ref{sec:app_train}.
\section{Experiments}
In this section, we introduce the experimental setup (\cref{sec:exp_setup}), experimental results (\cref{sec:exp_results}), and further analyses on our model (\Cref{sec:exp_general_if,sec:exp_ablation,sec:exp_category}).

\begin{table*}[ht]
\centering
\small{
\resizebox{\linewidth}{!}{
\begin{tabular}{llrrrrrrrrrrrr}
\toprule
\multirow{2}{*}{Model} & \multirow{2}{*}{Backbone} & \multicolumn{5}{c}{IFEval} & \multicolumn{6}{c}{FollowBench (HSR)} & \multirow{2}{*}{AVG}  \\ \cmidrule{3-13} 
 &  & [S]P & [S]I & [L]P & [L]I & AVG & L1 & L2 & L3 & L4 & L5 & AVG &   \\ \midrule
GPT-3.5* & GPT & $59.0$ & $68.5$ & $64.0$ & $73.6$ & $66.3$ & $80.3$ & $68.0$ & $68.6$ & $61.1$ & $53.2$ & $66.2$ & $66.3$ \\
GPT-4$\dagger$ & GPT & $76.9$ & $83.6$ & $79.3$ & $85.4$ & $81.3$ & $84.7$ & $76.1$ & $71.3$ & $74.5$ & $62.4$ & $73.8$ & $77.6$\\ \midrule
Vicuna-v1.5-13B$\dagger$ & Llama2 & $43.1$ & $53.6$ & $46.6$ & $58.0$ & $50.3$ & \boldsymbol{$71.2$} & \boldsymbol{$61.3$} & $48.3$ & $38.0$ & $33.1$ & \boldsymbol{$50.4$} & $50.4$ \\
WizardLM-v1.2-13B & Llama2 & $43.6$ & $54.4$ & $48.4$ & $59.1$ & $51.4$ & $61.3$ & $51.6$ & $43.3$ & $37.5$ & $29.9$ & $44.7$ & $48.1$ \\
Conifer\textsubscript{SFT}-13B$\dagger$ & Llama2 & $42.9$ & $53.0$ & $47.5$ & $57.4$ & $50.2$ & $60.5$ & $53.6$ & \underline{$48.4$} & $40.7$  & $31.7$ & $47.0$ & $48.6$ \\
Zephyr-beta-7B$\dagger$ & Mistral & $32.0$ & $46.8$ & $44.9$ & $58.0$ & $45.4$ & $57.6$ & $51.9$ & $41.9$ & $41.4$ & $31.4$ & $44.8$ & $45.1$ \\
Mistral\textsubscript{Instruct}-7B & Mistral & \underline{$48.6$} & \underline{$59.8$} & \underline{$53.2$} & \underline{$64.3$} & \underline{$56.5$} & $57.1$ & $51.5$ & $43.6$ & $41.5$ & $33.2$ & $45.4$ & $50.9$ \\ 
Suri\textsubscript{I-ORPO}-7B & Mistral & $47.3$ & $58.0$ & $51.4$ & $62.0$ & $54.7$ & $45.4$ & $41.4$ & $24.2$ & $18.6$ & $15.2$ & $29.0$ & $41.9$ \\ 
Conifer\textsubscript{SFT}-7B$\dagger$ & Mistral & $45.8$ & $57.1$ & $50.8$ & $62.0$ & $53.9$ & $54.3$ & $49.5$ & $49.3$ & $40.8$ & $30.5$ & $44.9$ & $49.4$\\
Conifer\textsubscript{DPO}-7B$\dagger$ & Mistral & $48.1$ & $59.1$ & $52.3$ & $63.3$ & $55.7$ & $60.3$ & $53.6$ & $48.0$ & \boldsymbol{$47.1$} & \boldsymbol{$41.0$} & \underline{$50.0$} & \underline{$52.9$}\\ 
\midrule
Llama3-8B & Llama3 & $25.7$ & $36.8$ & $28.1$ & $35.1$ & $31.4$ & $4.8$ & $8.7$ & $8.8$ & $6.0$ & $9.8$ & $7.6$ & $19.5$ \\ 
\rowcolor{gray!20}
\llamacrab & Llama3 & $39.4$ & $50.2$ & $43.8$ & $54.2$ & $46.9$  & $57.5$ & $44.9$ & $34.9$ & $25.2$ & $20.0$ & $36.5$ & $41.7$ \\
\rowcolor{gray!20}
\llamacrab + DPO & Llama3 & $40.3$ & $52.0$ & $47.7$ & $58.9$ & $49.7$ & $64.6$ & $49.0$ & $41.6$ & $35.8$ & \underline{$36.8$} & $45.5$ & $47.6$ \\
Mistral-7B & Mistral & $18.5$ & $30.8$ & $19.6$ & $31.9$ & $25.2$ & $14.3$ & $16.6$ & $8.3$ & $5.8$ & $5.5$ & $10.1$ & $17.7$  \\
\rowcolor{gray!20}
\mistralcrab & Mistral & $47.9$ & $57.3$ & $51.6$ & $61.2$ & $54.5$ & $63.9$ & \underline{$54.4$} & $40.1$ & $30.4$ & $27.9$ & $43.3$ & $48.9$ \\
\rowcolor{gray!20}
\mistralcrab + DPO & Mistral & \boldsymbol{$49.7$} & \boldsymbol{$61.5$} & \boldsymbol{$57.7$} & \boldsymbol{$68.5$} & \boldsymbol{$59.3$} & \underline{$66.1$} & $53.6$ & \boldsymbol{$53.4$} & \underline{$42.4$} & $31.7$ & $49.4$ & \boldsymbol{$54.4$}\\
\bottomrule
\end{tabular}
}
}
\caption{Experimental results (\%) of the LLMs on IFEval and FollowBench. In IFEval, ``[S]'' and ``[L]' denote strict and loose accuracy, ``P'' and ``I'' indicate the prompt and instruction level. 
In FollowBench, L1 (simplest) to L5 (hardest) denote different difficulty levels.
We highlight the highest and second-highest scores of open-source LLMs using \textbf{bold} font and \underline{underline}. $\dagger$ and 
 * means the results are from~\citet{sun2024conifer} and ~\citet{he2024complex}.}
\label{tab:main_exp}
\end{table*}

\begin{table}
\centering
\small{
\begin{tabular}{lrr}
\toprule
Model  &  LC WinRate & WinRate  \\ \midrule
GPT-3.5-turbo-0613$\dagger$  & $22.4$ & $14.1$ \\
GPT-4-0613$\dagger$   & $30.2$ & $15.8$ \\ \midrule
WizardLM-70B$\dagger$ & $17.6$ & $14.4$ \\
WizardLM-v1.2-13B$\dagger$  & $14.5$ & $12.0$  \\
Vicuna-v1.5-13B$\dagger$  & $10.5$ & $6.7$ \\ \midrule
Zephyr-beta-7B$\dagger$  & $13.2$ & $11.0$ \\
Conifer\textsubscript{DPO}-7B$\dagger$  & \underline{$17.1$} & \underline{$11.3$} \\
\rowcolor{gray!20}
\mistralcrab & $13.3$ & $7.9$ \\ 
\rowcolor{gray!20}
\mistralcrab + DPO & \boldsymbol{$18.1$} & \boldsymbol{$17.6$} \\ 
\quad \texttt{(vs.) Conifer\textsubscript{DPO} } & $60.6$ & $63.5$ \\
\bottomrule
\end{tabular}
\caption{Winning rate (\%) of the investigated LLMs on Alpaca-Eval 2.0~\citep{alpaca_eval}. 
``LC'' denotes length-controlled~\citep{dubois2024length}.
$\dagger$ means the results are sourced from the original leaderboard.
}
\label{tab:general_if}
}
\end{table}

\subsection{Experimental Setup}
\label{sec:exp_setup}
\paragraph{Backbone Models}

\looseness=-1
We adopt two widely-used open-source base models, Mistral 7B~\citep{jiang2023mistral} and Llama 3 8B~\citep{dubey2024llama}, as our backbone models for developing \llamacrab and \mistralcrab. Specifically, we employ \texttt{Mistral-7B-v0.3} and \texttt{Meta-Llama-3-8B}, downloaded from Hugging Face~\citep{wolf2019huggingface}.
During the SFT stage, we adopt a $5\times10^{-6}$ learning rate, $256$ batch size, and train the Mistral for $4$ epochs and Llama 3 for $3$ epochs. During the DPO optimization stage, we adopt $5\times10^{-7}$ learning rate, $64$ batch size, 
and $1$ training epoch with beta $0.01$ for Mistral and $3$ epochs with beta $0.1$ for Llama 3.


\paragraph{Baselines}
Our baselines include popular open-source and proprietary LLMs, divided into three main categories for comparison:
(1) Proprietary LLMs, including GPT-3.5~\citep{chatgpt} and GPT-4~\citep{gpt4}. 
(2) General instruction-tuning LLMs, including Vicuna-V1.5 13B~\citep{chiang2023vicuna}, trained on the 125k ShareGPT dataset, WizardLM-V1.2 13B~\citep{xu2023wizardlm}, trained on the 196k Evol-Instruct dataset, Zephyr beta 7B~\citep{tunstall2023zephyr}, trained with the UltraFeedback~\citep{cui2023ultrafeedback} dataset using the DPO objective~\citep{dpo}, and Mistral-Instruct 7B v0.3~\citep{jiang2023mistral}, which achieves leading performance on chat benchmarks based on Mistral 7B
(3) Models specifically optimized for complex instruction-following tasks, including Suri-I-ORPO~\citep{pham2024suri}, which is optimized for multi-constraint instruction-following tasks in long-form text generation, and the Conifer series~\citep{sun2024conifer}, which are trained on the data where the constraints, instructions, and responses are all generated using GPT-4.



\paragraph{Evaluation Datasets}
We use two widely-used and challenging complex instruction-following datasets IFEval~\citep{zhou2023instruction} and FollowBench~\citep{jiang-etal-2024-followbench} for evaluation.
IFEval consists of $541$ instructions that can be automatically validated using Python scripts. Each instruction contains $1$ to $3$ constraints, primarily focusing on strict lexical and formatting constraints.
FollowBench is a fine-grained, multi-constraint instruction-following benchmark and it categorizes the difficulty into five levels (L1 to L5) based on the number of constraints of an instruction, where L1 represents the simplest level with only one constraint, while L5 is the most difficult, with a combination of five constraints. It also includes five constraint categories, including \textit{content}, \textit{situation}, \textit{style}, \textit{format}, and \textit{example}, along with a \textit{mixed} constraint category that combines various categories of constraints. FollowBench contains a total of $820$ instructions across more than $50$ different NLP tasks, and it is automatically evaluated using either Python scripts or GPT-4. Please refer to the original paper for more details~\citep{jiang-etal-2024-followbench}.

\subsection{Experimental Results}
\label{sec:exp_results}


The experimental results are presented in Table~\ref{tab:main_exp}. Our observations are as follows: (1) After training on the \ourmodel dataset, our models significantly outperform the corresponding base models and the open-source models trained through SFT on general instruction-following datasets. Our DPO version of models achieves the best performance among the compared models. It demonstrates the effectiveness of our data and training approach. (2) Our models surpass Conifer~\citet{sun2024conifer}, which is specifically trained for complex instruction-following, on IFEval. It suggests that our model performs better in following lexical and format constraints. However, our models slightly lag behind Conifer on FollowBench. We provide an in-depth discussion on the performance across different constraint categories in FollowBench in \cref{sec:exp_category}. We observe that the performance decline is primarily due to the style constraint, where our models significantly underperform in this constraint category compared to Conifer. Nonetheless, our models achieve significant improvements in real-world scenarios, i.e., the \textit{mixed} constraint in FollowBench~\citep{jiang-etal-2024-followbench}, compared to Conifer. (3) Training with the DPO objective consistently improves model performance on both evaluation datasets. In this paper, we focus on constructing high-quality SFT data by constraint back-translation, we leave the development of DPO data for complex instruction-following as future work.


\subsection{Analysis on General Instruction Following}
\label{sec:exp_general_if}


\begin{figure*}[t]
    \includegraphics[width=1.0\linewidth]{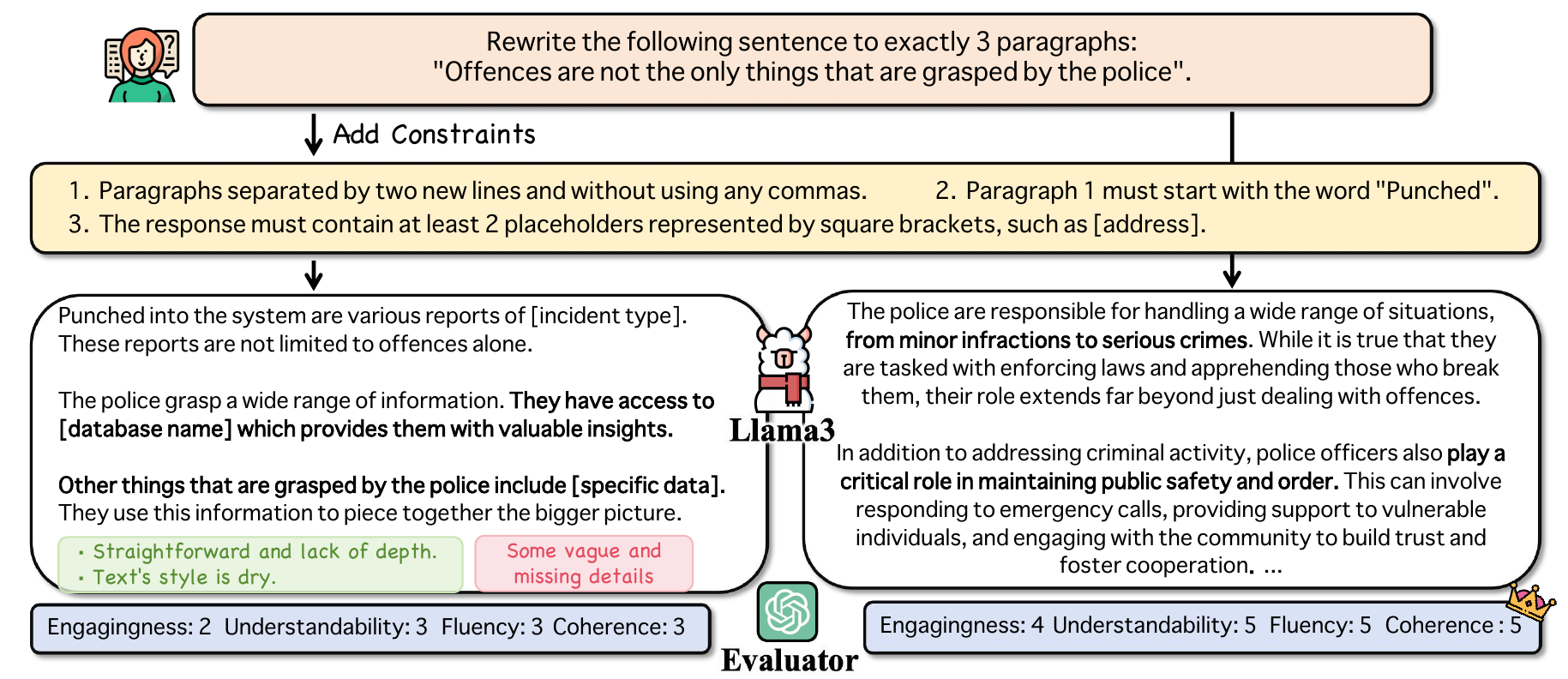} 
    \caption{An example of responses generated with and without constraints by Llama3-70B-Instruct. The evaluator is \texttt{gpt-4o-0806}. For better visualization, we present only a subset of the responses generated without constraints.
    }
    \label{fig:case}
\end{figure*}

The complex instruction-following ability not only involves following complex constraints but also encompasses the basic ability to follow instructions themselves, e.g., ``\textit{Write a blog on French cuisine}'', named as general instruction following. In this section, we further evaluate our model's general instruction-following capability. Given that IFEval and FollowBench primarily focus on evaluating the ability to follow constraints, we adopt another widely-used dataset,  AlpacaEval~\citep{alpaca_eval}, which serves as an easy-to-use and high-quality automatic evaluator for instruction-following ability. Specifically, we use AlpacaEval 2.0, which contains $805$ instructions, and use \texttt{gpt-4-1106-preview} as the evaluator to get the final weighted win rate. The evaluation results are presented in Table~\ref{tab:general_if}, where the ``LC WinRate'' represents the length-controlled win rate~\citep{dubois2024length}. The default reference model is \texttt{gpt-4-1106-preview}. We can observe that our model significantly outperforms the baseline model Conifer and even exceeds the performance of the 70B version of WizardLM. We also conduct a head-to-head comparison between our model and Conifer. The LC win rate of our model reaches $60.6$, significantly outperforming Conifer, which demonstrates that our model possesses a superior general instruction-following capability.

\begin{figure}[t]
    \includegraphics[width=1.0\linewidth]{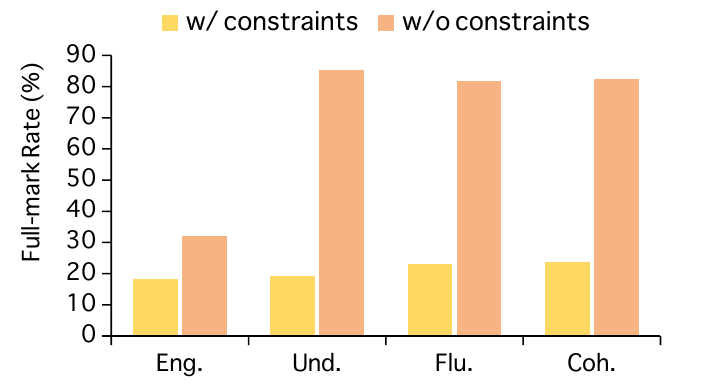} 
    \caption{
    Full-mark rates (\%) of the responses generated with and without constraints. The evaluator is \texttt{gpt-4o-0806}, focusing on four widely-used dimensions: Engagingness (Eng.), Understandability (Und.), Fluency (Flu.), and Coherence (Coh.).
    }
    \label{fig:prop_perfect_score}
\end{figure}

We conduct a further analysis to explore the potential reasons why our model outperforms Conifer in general instruction-following. The primary difference between our model and Conifer is the data construction process. We utilize \textit{constraint back-translation}, where the response is generated directly from the instruction without constraints. In contrast, Conifer uses instruction and corresponding constraints to generate the response. We hypothesize that a possible reason is that the response quality in Conifer is lower than \ourmodel, that is, generating a response conditioned on both instruction and constraints may result in \textbf{lower content quality}, such as lower coherence, compared to the response directly generated from instruction without constraints.
Intuitively, incorporating constraints may limit the model's capacity when generating responses. To validate this intuition, we conduct a controlled analytical experiment. Specifically, we sample $100$ instructions and their corresponding constraints from IFEval and FollowBench. We first use Llama3-70B-Instruct to generate responses based only on the instructions (w/o constraints). Then, we include the additional constraints and generate corresponding responses (w/ constraints). Following previous work on automated evaluation using advanced LLMs~\citep{bai2024benchmarking, chanchateval}, we employ \texttt{gpt-4o-0806} as the evaluator, assessing the responses on four dimensions: \textit{engagingness}, \textit{understandability}, \textit{fluency}, and \textit{coherence}, with scores ranging from $1$ to $5$. We report the full-mark (score $5$) rate for each dimension. The results are shown in Figure~\ref{fig:prop_perfect_score}.
We can observe that responses generated with constraints significantly underperform those generated without constraints, which suggests that involving constraints when generation may reduce the content quality of the final response. 
We further conduct a case study, illustrated in Figure~\ref{fig:case}, showing an example of responses generated with and without constraints. We can find that when involving constraints, the response includes vague terms, such as ``\textit{[database name]}'', and lacks sufficient details and depth. While previous work on complex instruction-following mainly focuses on enhancing the ability to follow multiple constraints, we encourage future work to prioritize response content quality, and constraint back-translation can serve as a potential solution.
Assessment details of this analysis are in appendix~\ref{sec:app_pilot}.

\subsection{Ablation Study}
\label{sec:exp_ablation}
\begin{table}
\centering
\small{
\resizebox{\linewidth}{!}{
\begin{tabular}{lcccc}
\toprule
\multirow{2}{*}{Model} & IFEval & \multicolumn{2}{c}{FollowBench} & \multirow{2}{*}{AVG} \\ \cmidrule{3-4}
 & AVG & L1-L2 & L3-L5 & \\ \midrule
 \mistralcrab & $54.5$ & $59.1$ & $32.8$ & $48.9$  \\ \midrule
\quad \texttt{(-)} \textit{Reverse training} & $52.1$ & $56.2$ & $33.5$ & $47.3$ \\
\quad \texttt{(-)} \textit{Forward training} & $53.9$ & $57.1$ & $32.1$ & $48.0$ \\
\quad \texttt{(-)} \textit{In-Context Demons} & $53.6$ & $55.8$ & $30.0$ & $47.0$  \\ 
\midrule
InstBackT\textsubscript{SFT} & $52.7$ & $55.4$ & $29.3$ & $46.2$ \\
\bottomrule
\end{tabular}
}
\caption{Experimental results (\%) of the ablation study. \textit{In-Context Demons} denotes in-context demonstrations.
}
\label{tab:ablation}
}
\end{table}

We conduct an ablation study to analyze the key factors influencing model performance. Specifically, we investigate three key factors in developing our model: \textit{reverse training}, \textit{forward training}, i.e., standard supervised fine-tuning, and \textit{in-context demonstrations}. We exclude each factor and keep all other conditions identical, to the model separately. 
When excluding reverse and forward training, we set the loss ratio $\alpha$ in \cref{sec:model_training} to $1$ and $0$, respectively.
The backbone model is Mistral. 
The results are presented in Table~\ref{tab:ablation}, where L1-L2 in FollowBench represent simpler constraints and L3-L5 denote more complex constraints.
We can observe that removing any of these factors leads to a decline in model performance, which demonstrates the effectiveness of these factors in developing our model.
For more complex constraints following, adding in-context demonstrations during training is effective, as excluding in-context demonstrations leads to a significant performance drop in L3-L5.
The reason may be that in-context demonstrations enhance the model's ability to understand multiple in-context instructions and complex constraints.

We further compare with a competitive baseline model, InstBackT\textsubscript{SFT}, which is trained on the data generated by \textit{instruction back-translation}~\citep{liself2024}.
The key difference between instruction and constraint back-translation is that the former uses advanced LLMs to generate both instructions and constraints from responses, while the latter focuses on generating constraints from instruction and response pairs. 
The results are presented in Table~\ref{tab:ablation}.
We can observe that InstBackT\textsubscript{SFT} significantly underperforms compared to \mistralcrab, which suggests that instruction back-translation may produce lower-quality data for complex instruction following. The possible reason is that generating both instructions and constraints simultaneously is more challenging than generating constraints alone. It further demonstrates the efficacy of the constraint back-translation method in creating high-quality training data for complex instruction following.


\subsection{Analysis on Constraint Category}
\label{sec:exp_category}
\label{sec:category_analysis}

\begin{figure}[t]
    \includegraphics[width=1.0\linewidth]{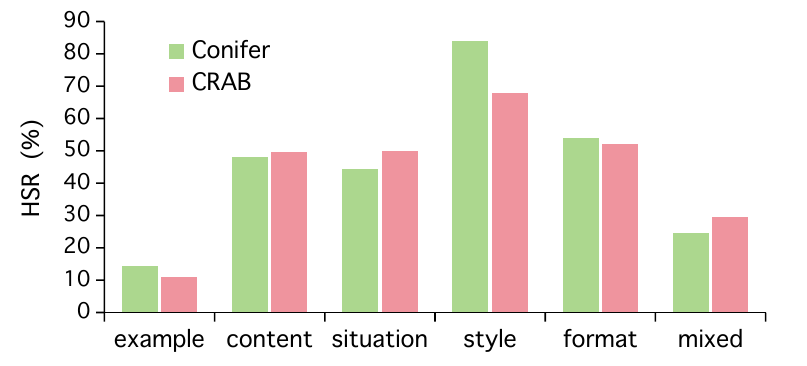} 
    \caption{
    Experimental results on different categories of constraints in the FollowBench of
    \mistralcrab and Conifer\textsubscript{SFT}.
    }
    \label{fig:followbench}
\end{figure}

We further investigate our model's performance across different constraint categories to analyze its strengths and potential limitations. Specifically, we analyze the results on FollowBench, which includes five categories of constraints, including \textit{example}, \textit{content}, \textit{situation}, \textit{style}, and \textit{format}. 
Please refer to the original paper~\citep{jiang-etal-2024-followbench} for the detailed definitions for each constraint category. FollowBench also includes a \textit{mixed} category which is designed for simulating \textbf{real-world} scenarios~\citep{jiang-etal-2024-followbench}, where various types of constraints are combined to form the final constraint. We compare our model \mistralcrab with the Conifer model, which is trained on the data generated using the standard pipeline: generating the constraints first and then generating the response based on the instruction and constraints.
The results on different constraint categories of FollowBench are shown in Figure~\ref{fig:followbench}. We can observe that our model significantly outperforms Conifer on the \textit{mixed} constraint, which represents real-world scenarios, suggesting that our model is more effective in handling complex instruction-following scenarios.
However, in the \textit{style} constraint category, e.g., ``\textit{Write in the style of Shakespeare}'', our model performs significantly worse than Conifer. The possible reason is that the style constraints in our dataset \ourmodel may be not sufficiently diverse. The data for \textit{style} constraints requires deliberate construction, and the pipeline that generates constraints first and then responses is more effective at generating diverse style constraints, 
but the responses in our seed data have limited style diversity.
It suggests a limitation of constraint back-translation, as it relies on diverse responses to generate specific categories of constraints, such as style constraint. 
Combining the constraint back-translation method with other data generation methods to produce higher-quality data for those specific constraints can further enhance the model's complex instruction-following ability, and we leave this exploration as future work.

\section{Related Work}

\subsection{Instruction Following}
\looseness=-1
Instruction following task involves following user intentions to generate helpful responses, which is fundamental to modern LLMs~\citep{zhang2023instruction}. \citet{ouyang2022training} first propose the practice of aligning LLMs to follow human instructions, using SFT and RLHF to train models, which is the key factor in the success of ChatGPT~\citep{chatgpt}. Subsequently, numerous studies focus on enhancing the instruction-following capabilities of LLMs, particularly for open-source models, which can be summarized in two main aspects: (1) data-driven approaches, which design an automated pipeline or use human annotation to produce high-quality training data~\citep{xu2023wizardlm,taori2023alpaca,alpaca-gpt4,chiang2023vicuna,mukherjee2023orca, qi2024adelie,liu2024aligning,liself2024,bai2024longalign, Peng2024PretrainingDF}. (2) new training methods, including novel objectives~\citep{dpo,gallego2024refined,zhou2024beyond,hejna2024inverse,meng2024simpo} or training pipelines~\citep{tunstall2023zephyr,liself2024,yuanself2024,chenself2024}.


\looseness=-1
A more challenging instruction following scenario is complex constrained instruction following, where the responses should further satisfy specific constraints, such as length, keyword, and format. Previous studies have shown that LLMs struggle to follow these instructions~\citep{jiang-etal-2024-followbench, qin2024infobench, chen2024sifo}. Recent efforts focus on enhancing this ability by constructing high-quality training data~\citep{sun2024conifer,he2024complex}. This process typically involves collecting a set of instructions, constructing constraints, and then generating responses based on the instructions and constraints using advanced LLMs. This work introduces \textit{constraint back-translation}, which generates constraints from instruction-response pairs, reducing data construction costs and noise.

\subsection{Back-translation}
Back-translation is first proposed in the field of machine translation~\citep{sennrich2015improving,hoang2018iterative}, which mainly is used for data augmentation.
It first trains a model to back-translate the target language into the source language, then uses this model to generate parallel training data from a large amount of monolingual target language data, which sufficiently saves human translation efforts. 
Considering its simplicity and efficacy, back-translation has also been widely applied to various tasks, such as style transfer~\citep{prabhumoye2018style,toshevska2021review} and paraphrase generation~\citep{wieting2017learning,mallinson2017paraphrasing}.


Recently, several studies have explored applying back-translation to the field of large language models to efficiently generate high-quality data automatically~\citep{li2023self,pham2024suri,koksal2023longform}.
\citet{li2023self} proposed reversing the training objective to automatically generate corresponding instructions for existing unsupervised corpora, while \citet{pham2024suri} and \citet{koksal2023longform} leveraged the powerful general capabilities of LLMs to generate instructions from the corpus directly. Although \citet{pham2024suri} also generated constraints, it fell within the realm of \textit{instruction back-translation} and did not involve dedicated optimization or exploration for constraint generation.
In this work, we propose \textit{constraint back-translation}, an effective data generation approach that generates high-quality constraints based on instruction-response pairs.



\section{Conclusion}

In this paper, we aim to enhance large language models' capability for complex constrained instruction following. We propose a \textit{constraint back-translation} data generation method, which can reduce data noise and generation costs, resulting in a high-quality complex instruction-following dataset \ourmodel. We also propose a \textit{reverse training} method and develop \llamacrab and \mistralcrab based on \ourmodel. Extensive experiments demonstrate the effectiveness of our data generation and training methods. 
We further conduct extensive analytical experiments and discuss the key factors, advantages, and potential limitations of our model.

\section*{Limitations}
%

As discussed in \cref{sec:exp_category}, for certain types of constraints, such as style constraint, the constraints generated through constraint back-translation may lack sufficient diversity if the original response data itself is not diverse enough.
We leave further improvements to constraint back-translation as future work. Another limitation of our study is that we do not use a larger base model due to computational constraints. We believe that using a larger base model could develop a more advanced LLM in following complex constraints, but this does not affect our overall experimental conclusions.


\section*{Ethical Considerations}
We discuss potential ethical concerns related to this work:
(1) \textbf{Intellectual property}. Our research leverages several widely used SFT datasets, and we strictly comply with their licensing terms. We will share \ourdata under the CC BY-SA 4.0 license\footnote{\url{https://creativecommons.org/licenses/by-sa/4.0/}}.
(2) \textbf{Intended use and Potential risk control}.  The goal of this paper is to introduce \ourdata, designed to enhance the performance of LLMs on complex instruction tasks. \ourdata is built using widely available public datasets. We trust that the original publishers have anonymized and sanitized these datasets appropriately. The data construction process does not include additional social bias.
Additionally, we randomly sampled $100$ instances from our dataset and found no sensitive information.
(3) \textbf{AI assistance}. We used GPT-4 to paraphrase some sentences and check grammar.

\bibliography{custom}

\begin{thebibliography}{64}
\providecommand{\natexlab}[1]{#1}

\bibitem[{Amini et~al.(2019)Amini, Gabriel, Lin, Koncel-Kedziorski, Choi, and Hajishirzi}]{amini2019mathqa}
Aida Amini, Saadia Gabriel, Peter Lin, Rik Koncel-Kedziorski, Yejin Choi, and Hannaneh Hajishirzi. 2019.
\newblock \href {https://arxiv.org/abs/1905.13319} {Mathqa: Towards interpretable math word problem solving with operation-based formalisms}.
\newblock \emph{Preprint}, arXiv:1905.13319.

\bibitem[{Bai et~al.(2024{\natexlab{a}})Bai, Lv, Zhang, He, Qi, Hou, Tang, Dong, and Li}]{bai2024longalign}
Yushi Bai, Xin Lv, Jiajie Zhang, Yuze He, Ji~Qi, Lei Hou, Jie Tang, Yuxiao Dong, and Juanzi Li. 2024{\natexlab{a}}.
\newblock Longalign: A recipe for long context alignment of large language models.
\newblock \emph{arXiv preprint arXiv:2401.18058}.

\bibitem[{Bai et~al.(2024{\natexlab{b}})Bai, Ying, Cao, Lv, He, Wang, Yu, Zeng, Xiao, Lyu et~al.}]{bai2024benchmarking}
Yushi Bai, Jiahao Ying, Yixin Cao, Xin Lv, Yuze He, Xiaozhi Wang, Jifan Yu, Kaisheng Zeng, Yijia Xiao, Haozhe Lyu, et~al. 2024{\natexlab{b}}.
\newblock Benchmarking foundation models with language-model-as-an-examiner.
\newblock \emph{Advances in NeurlPS}, 36.

\bibitem[{Berglund et~al.(2023)Berglund, Tong, Kaufmann, Balesni, Stickland, Korbak, and Evans}]{berglund2023reversal}
Lukas Berglund, Meg Tong, Max Kaufmann, Mikita Balesni, Asa~Cooper Stickland, Tomasz Korbak, and Owain Evans. 2023.
\newblock The reversal curse: Llms trained on" a is b" fail to learn" b is a".
\newblock \emph{arXiv preprint arXiv:2309.12288}.

\bibitem[{Campos et~al.(2020)Campos, Mangaravite, Pasquali, Jorge, Nunes, and Jatowt}]{campos2020yake}
Ricardo Campos, V{\'\i}tor Mangaravite, Arian Pasquali, Al{\'\i}pio Jorge, C{\'e}lia Nunes, and Adam Jatowt. 2020.
\newblock Yake! keyword extraction from single documents using multiple local features.
\newblock \emph{Information Sciences}, 509:257--289.

\bibitem[{Chan et~al.(2024)Chan, Chen, Su, Yu, Xue, Zhang, Fu, and Liu}]{chanchateval}
Chi-Min Chan, Weize Chen, Yusheng Su, Jianxuan Yu, Wei Xue, Shanghang Zhang, Jie Fu, and Zhiyuan Liu. 2024.
\newblock Chateval: Towards better llm-based evaluators through multi-agent debate.
\newblock In \emph{ICLR}.

\bibitem[{Chen et~al.(2024{\natexlab{a}})Chen, Liao, Qi, Eustratiadis, Monz, Bisazza, and de~Rijke}]{chen2024sifo}
Xinyi Chen, Baohao Liao, Jirui Qi, Panagiotis Eustratiadis, Christof Monz, Arianna Bisazza, and Maarten de~Rijke. 2024{\natexlab{a}}.
\newblock The sifo benchmark: Investigating the sequential instruction following ability of large language models.
\newblock \emph{arXiv preprint arXiv:2406.19999}.

\bibitem[{Chen et~al.(2024{\natexlab{b}})Chen, Deng, Yuan, Ji, and Gu}]{chenself2024}
Zixiang Chen, Yihe Deng, Huizhuo Yuan, Kaixuan Ji, and Quanquan Gu. 2024{\natexlab{b}}.
\newblock Self-play fine-tuning converts weak language models to strong language models.
\newblock In \emph{ICML}.

\bibitem[{Chiang et~al.(2023)Chiang, Li, Lin, Sheng, Wu, Zhang, Zheng, Zhuang, Zhuang, Gonzalez et~al.}]{chiang2023vicuna}
Wei-Lin Chiang, Zhuohan Li, Zi~Lin, Ying Sheng, Zhanghao Wu, Hao Zhang, Lianmin Zheng, Siyuan Zhuang, Yonghao Zhuang, Joseph~E Gonzalez, et~al. 2023.
\newblock Vicuna: An open-source chatbot impressing gpt-4 with 90\%* chatgpt quality.
\newblock \emph{See https://vicuna. lmsys. org (accessed 14 April 2023)}, 2(3):6.

\bibitem[{Clark et~al.(2018)Clark, Cowhey, Etzioni, Khot, Sabharwal, Schoenick, and Tafjord}]{clark2018think}
Peter Clark, Isaac Cowhey, Oren Etzioni, Tushar Khot, Ashish Sabharwal, Carissa Schoenick, and Oyvind Tafjord. 2018.
\newblock \href {https://arxiv.org/abs/1803.05457} {Think you have solved question answering? {T}ry arc, the ai2 reasoning challenge}.
\newblock \emph{ArXiv preprint}.

\bibitem[{Cobbe et~al.(2021)Cobbe, Kosaraju, Bavarian, Hilton, Nakano, Hesse, and Schulman}]{cobbe2021training}
Karl Cobbe, Vineet Kosaraju, Mohammad Bavarian, Jacob Hilton, Reiichiro Nakano, Christopher Hesse, and John Schulman. 2021.
\newblock \href {https://arxiv.org/abs/2110.14168} {Training verifiers to solve math word problems}.
\newblock \emph{Preprint}, arXiv:2110.14168.

\bibitem[{Cui et~al.(2023)Cui, Yuan, Ding, Yao, Zhu, Ni, Xie, Liu, and Sun}]{cui2023ultrafeedback}
Ganqu Cui, Lifan Yuan, Ning Ding, Guanming Yao, Wei Zhu, Yuan Ni, Guotong Xie, Zhiyuan Liu, and Maosong Sun. 2023.
\newblock Ultrafeedback: Boosting language models with high-quality feedback.
\newblock \emph{arXiv preprint arXiv:2310.01377}.

\bibitem[{Dong et~al.(2024)Dong, Lu, Li, Xia, Yu, Zhou, and Zhou}]{dong2024self}
Guanting Dong, Keming Lu, Chengpeng Li, Tingyu Xia, Bowen Yu, Chang Zhou, and Jingren Zhou. 2024.
\newblock Self-play with execution feedback: Improving instruction-following capabilities of large language models.
\newblock \emph{arXiv preprint arXiv:2406.13542}.

\bibitem[{Dubey et~al.(2024)Dubey, Jauhri, Pandey, Kadian, Al-Dahle, Letman, Mathur, Schelten, Yang, Fan et~al.}]{dubey2024llama}
Abhimanyu Dubey, Abhinav Jauhri, Abhinav Pandey, Abhishek Kadian, Ahmad Al-Dahle, Aiesha Letman, Akhil Mathur, Alan Schelten, Amy Yang, Angela Fan, et~al. 2024.
\newblock The llama 3 herd of models.
\newblock \emph{arXiv preprint arXiv:2407.21783}.

\bibitem[{Dubois et~al.(2024)Dubois, Galambosi, Liang, and Hashimoto}]{dubois2024length}
Yann Dubois, Bal{\'a}zs Galambosi, Percy Liang, and Tatsunori~B Hashimoto. 2024.
\newblock Length-controlled alpacaeval: A simple way to debias automatic evaluators.
\newblock \emph{arXiv preprint arXiv:2404.04475}.

\bibitem[{Es(2023)}]{Orca-Chat}
Shahul Es. 2023.
\newblock Orca-chat: A high-quality explanation-style chat dataset.

\bibitem[{Gallego(2024)}]{gallego2024refined}
V{\'\i}ctor Gallego. 2024.
\newblock Refined direct preference optimization with synthetic data for behavioral alignment of llms.
\newblock \emph{arXiv preprint arXiv:2402.08005}.

\bibitem[{Golovneva et~al.(2024)Golovneva, Allen-Zhu, Weston, and Sukhbaatar}]{golovneva2024reverse}
Olga Golovneva, Zeyuan Allen-Zhu, Jason Weston, and Sainbayar Sukhbaatar. 2024.
\newblock Reverse training to nurse the reversal curse.
\newblock \emph{arXiv preprint arXiv:2403.13799}.

\bibitem[{He et~al.(2024)He, Zeng, He, Liang, and Xiao}]{he2024complex}
Qianyu He, Jie Zeng, Qianxi He, Jiaqing Liang, and Yanghua Xiao. 2024.
\newblock \href {https://arxiv.org/abs/2404.15846} {From complex to simple: Enhancing multi-constraint complex instruction following ability of large language models}.
\newblock \emph{Preprint}, arXiv:2404.15846.

\bibitem[{Hejna and Sadigh(2024)}]{hejna2024inverse}
Joey Hejna and Dorsa Sadigh. 2024.
\newblock Inverse preference learning: Preference-based rl without a reward function.
\newblock \emph{Advances in Neural Information Processing Systems}, 36.

\bibitem[{Hendrycks et~al.(2021)Hendrycks, Burns, Basart, Zou, Mazeika, Song, and Steinhardt}]{Hendrycks2020MeasuringMM}
Dan Hendrycks, Collin Burns, Steven Basart, Andy Zou, Mantas Mazeika, Dawn Song, and Jacob Steinhardt. 2021.
\newblock \href {https://openreview.net/forum?id=d7KBjmI3GmQ} {Measuring massive multitask language understanding}.
\newblock In \emph{Proceedings of ICLR}.

\bibitem[{Hoang et~al.(2018)Hoang, Koehn, Haffari, and Cohn}]{hoang2018iterative}
Cong Duy~Vu Hoang, Philipp Koehn, Gholamreza Haffari, and Trevor Cohn. 2018.
\newblock Iterative back-translation for neural machine translation.
\newblock In \emph{2nd Workshop on Neural Machine Translation and Generation}, pages 18--24. ACL.

\bibitem[{Ivison et~al.(2023)Ivison, Wang, Pyatkin, Lambert, Peters, Dasigi, Jang, Wadden, Smith, Beltagy et~al.}]{ivison2023camels}
Hamish Ivison, Yizhong Wang, Valentina Pyatkin, Nathan Lambert, Matthew Peters, Pradeep Dasigi, Joel Jang, David Wadden, Noah~A Smith, Iz~Beltagy, et~al. 2023.
\newblock Camels in a changing climate: Enhancing lm adaptation with tulu 2.
\newblock \emph{arXiv preprint arXiv:2311.10702}.

\bibitem[{Jiang et~al.(2023)Jiang, Sablayrolles, Mensch, Bamford, Chaplot, Casas, Bressand, Lengyel, Lample, Saulnier et~al.}]{jiang2023mistral}
Albert~Q Jiang, Alexandre Sablayrolles, Arthur Mensch, Chris Bamford, Devendra~Singh Chaplot, Diego de~las Casas, Florian Bressand, Gianna Lengyel, Guillaume Lample, Lucile Saulnier, et~al. 2023.
\newblock Mistral 7b.
\newblock \emph{arXiv preprint arXiv:2310.06825}.

\bibitem[{Jiang et~al.(2024)Jiang, Wang, Zeng, Zhong, Li, Mi, Shang, Jiang, Liu, and Wang}]{jiang-etal-2024-followbench}
Yuxin Jiang, Yufei Wang, Xingshan Zeng, Wanjun Zhong, Liangyou Li, Fei Mi, Lifeng Shang, Xin Jiang, Qun Liu, and Wei Wang. 2024.
\newblock \href {https://aclanthology.org/2024.acl-long.257} {{F}ollow{B}ench: A multi-level fine-grained constraints following benchmark for large language models}.
\newblock In \emph{Proceedings of ACL}, pages 4667--4688.

\bibitem[{K{\"o}ksal et~al.(2023)K{\"o}ksal, Schick, Korhonen, and Sch{\"u}tze}]{koksal2023longform}
Abdullatif K{\"o}ksal, Timo Schick, Anna Korhonen, and Hinrich Sch{\"u}tze. 2023.
\newblock Longform: Effective instruction tuning with reverse instructions.
\newblock \emph{arXiv preprint arXiv:2304.08460}.

\bibitem[{K{\"o}pf et~al.(2024)K{\"o}pf, Kilcher, von R{\"u}tte, Anagnostidis, Tam, Stevens, Barhoum, Nguyen, Stanley, Nagyfi et~al.}]{kopf2024openassistant}
Andreas K{\"o}pf, Yannic Kilcher, Dimitri von R{\"u}tte, Sotiris Anagnostidis, Zhi~Rui Tam, Keith Stevens, Abdullah Barhoum, Duc Nguyen, Oliver Stanley, Rich{\'a}rd Nagyfi, et~al. 2024.
\newblock Openassistant conversations-democratizing large language model alignment.
\newblock \emph{Advances in NeurlPS}, 36.

\bibitem[{Li et~al.(2023{\natexlab{a}})Li, Yu, Zhou, Schick, Levy, Zettlemoyer, Weston, and Lewis}]{li2023self}
Xian Li, Ping Yu, Chunting Zhou, Timo Schick, Omer Levy, Luke Zettlemoyer, Jason Weston, and Mike Lewis. 2023{\natexlab{a}}.
\newblock Self-alignment with instruction backtranslation.
\newblock \emph{arXiv preprint arXiv:2308.06259}.

\bibitem[{Li et~al.(2024)Li, Yu, Zhou, Schick, Levy, Zettlemoyer, Weston, and Lewis}]{liself2024}
Xian Li, Ping Yu, Chunting Zhou, Timo Schick, Omer Levy, Luke Zettlemoyer, Jason~E Weston, and Mike Lewis. 2024.
\newblock Self-alignment with instruction backtranslation.
\newblock In \emph{ICLR}.

\bibitem[{Li et~al.(2023{\natexlab{b}})Li, Zhang, Dubois, Taori, Gulrajani, Guestrin, Liang, and Hashimoto}]{alpaca_eval}
Xuechen Li, Tianyi Zhang, Yann Dubois, Rohan Taori, Ishaan Gulrajani, Carlos Guestrin, Percy Liang, and Tatsunori~B. Hashimoto. 2023{\natexlab{b}}.
\newblock Alpacaeval: An automatic evaluator of instruction-following models.
\newblock \url{https://github.com/tatsu-lab/alpaca_eval}.

\bibitem[{Lin(2004)}]{lin2004rouge}
Chin-Yew Lin. 2004.
\newblock Rouge: A package for automatic evaluation of summaries.
\newblock In \emph{Text summarization branches out}, pages 74--81.

\bibitem[{Liu et~al.(2024)Liu, Zhang, Yao, Cao, Hou, and Li}]{liu2024aligning}
Yantao Liu, Zhao Zhang, Zijun Yao, Shulin Cao, Lei Hou, and Juanzi Li. 2024.
\newblock Aligning teacher with student preferences for tailored training data generation.
\newblock \emph{arXiv preprint arXiv:2406.19227}.

\bibitem[{Loper and Bird(2002)}]{loper2002nltk}
Edward Loper and Steven Bird. 2002.
\newblock Nltk: The natural language toolkit.
\newblock \emph{arXiv preprint cs/0205028}.

\bibitem[{Mallinson et~al.(2017)Mallinson, Sennrich, and Lapata}]{mallinson2017paraphrasing}
Jonathan Mallinson, Rico Sennrich, and Mirella Lapata. 2017.
\newblock Paraphrasing revisited with neural machine translation.
\newblock In \emph{Proceedings of ACL}, pages 881--893.

\bibitem[{Meng et~al.(2024)Meng, Xia, and Chen}]{meng2024simpo}
Yu~Meng, Mengzhou Xia, and Danqi Chen. 2024.
\newblock Simpo: Simple preference optimization with a reference-free reward.
\newblock \emph{arXiv preprint arXiv:2405.14734}.

\bibitem[{Mukherjee et~al.(2023)Mukherjee, Mitra, Jawahar, Agarwal, Palangi, and Awadallah}]{mukherjee2023orca}
Subhabrata Mukherjee, Arindam Mitra, Ganesh Jawahar, Sahaj Agarwal, Hamid Palangi, and Ahmed Awadallah. 2023.
\newblock Orca: Progressive learning from complex explanation traces of gpt-4.
\newblock \emph{arXiv preprint arXiv:2306.02707}.

\bibitem[{OpenAI(2022)}]{chatgpt}
OpenAI. 2022.
\newblock \href {https://openai.com/blog/chatgpt} {Introducing {C}hat{GPT}}.

\bibitem[{OpenAI(2024)}]{gpt4}
OpenAI. 2024.
\newblock \href {https://arxiv.org/abs/2303.08774} {Gpt-4 technical report}.
\newblock \emph{Preprint}, arXiv:2303.08774.

\bibitem[{Ouyang et~al.(2022)Ouyang, Wu, Jiang, Almeida, Wainwright, Mishkin, Zhang, Agarwal, Slama, Ray et~al.}]{ouyang2022training}
Long Ouyang, Jeffrey Wu, Xu~Jiang, Diogo Almeida, Carroll Wainwright, Pamela Mishkin, Chong Zhang, Sandhini Agarwal, Katarina Slama, Alex Ray, et~al. 2022.
\newblock Training language models to follow instructions with human feedback.
\newblock \emph{Advances in NeurlPS}, 35:27730--27744.

\bibitem[{Peng et~al.(2023)Peng, Li, He, Galley, and Gao}]{alpaca-gpt4}
Baolin Peng, Chunyuan Li, Pengcheng He, Michel Galley, and Jianfeng Gao. 2023.
\newblock Instruction tuning with gpt-4.
\newblock \emph{arXiv preprint arXiv:2304.03277}.

\bibitem[{Peng et~al.(2024)Peng, Lv, Bai, Yao, Zhang, Hou, and Li}]{Peng2024PretrainingDF}
Hao Peng, Xin Lv, Yushi Bai, Zijun Yao, Jiajie Zhang, Lei Hou, and Juanzi Li. 2024.
\newblock \href {https://api.semanticscholar.org/CorpusID:273507110} {Pre-training distillation for large language models: A design space exploration}.
\newblock \emph{ArXiv}, abs/2410.16215.

\bibitem[{Pham et~al.(2024)Pham, Sun, and Iyyer}]{pham2024suri}
Chau~Minh Pham, Simeng Sun, and Mohit Iyyer. 2024.
\newblock Suri: Multi-constraint instruction following for long-form text generation.
\newblock \emph{arXiv preprint arXiv:2406.19371}.

\bibitem[{Prabhumoye et~al.(2018)Prabhumoye, Tsvetkov, Salakhutdinov, and Black}]{prabhumoye2018style}
Shrimai Prabhumoye, Yulia Tsvetkov, Ruslan Salakhutdinov, and Alan~W Black. 2018.
\newblock Style transfer through back-translation.
\newblock \emph{arXiv preprint arXiv:1804.09000}.

\bibitem[{Qi et~al.(2024)Qi, Peng, Wang, Xu, Hou, and Li}]{qi2024adelie}
Yunjia Qi, Hao Peng, Xiaozhi Wang, Bin Xu, Lei Hou, and Juanzi Li. 2024.
\newblock Adelie: Aligning large language models on information extraction.
\newblock \emph{arXiv preprint arXiv:2405.05008}.

\bibitem[{Qin et~al.(2024)Qin, Song, Hu, Yao, Cho, Wang, Wu, Liu, Liu, and Yu}]{qin2024infobench}
Yiwei Qin, Kaiqiang Song, Yebowen Hu, Wenlin Yao, Sangwoo Cho, Xiaoyang Wang, Xuansheng Wu, Fei Liu, Pengfei Liu, and Dong Yu. 2024.
\newblock Infobench: Evaluating instruction following ability in large language models.
\newblock \emph{arXiv preprint arXiv:2401.03601}.

\bibitem[{Rafailov et~al.(2023)Rafailov, Sharma, Mitchell, Manning, Ermon, and Finn}]{dpo}
Rafael Rafailov, Archit Sharma, Eric Mitchell, Christopher~D. Manning, Stefano Ermon, and Chelsea Finn. 2023.
\newblock \href {http://papers.nips.cc/paper\_files/paper/2023/hash/a85b405ed65c6477a4fe8302b5e06ce7-Abstract-Conference.html} {Direct preference optimization: Your language model is secretly a reward model}.
\newblock In \emph{Advances in {NeurIPs}}.

\bibitem[{Sap et~al.(2019)Sap, Rashkin, Chen, Le~Bras, and Choi}]{sap2019social}
Maarten Sap, Hannah Rashkin, Derek Chen, Ronan Le~Bras, and Yejin Choi. 2019.
\newblock \href {https://doi.org/10.18653/v1/D19-1454} {Social {IQ}a: Commonsense reasoning about social interactions}.
\newblock In \emph{Proceedings of EMNLP}, pages 4463--4473.

\bibitem[{Sennrich(2015)}]{sennrich2015improving}
Rico Sennrich. 2015.
\newblock Improving neural machine translation models with monolingual data.
\newblock \emph{arXiv preprint arXiv:1511.06709}.

\bibitem[{Sun et~al.(2024)Sun, Liu, Li, Wang, Dong, Lin, and Huang}]{sun2024conifer}
Haoran Sun, Lixin Liu, Junjie Li, Fengyu Wang, Baohua Dong, Ran Lin, and Ruohui Huang. 2024.
\newblock \href {https://arxiv.org/abs/2404.02823} {Conifer: Improving complex constrained instruction-following ability of large language models}.
\newblock \emph{arxiv preprint arXiv:2404.02823}.

\bibitem[{Taori et~al.(2023)Taori, Gulrajani, Zhang, Dubois, Li, Guestrin, Liang, and Hashimoto}]{taori2023alpaca}
Rohan Taori, Ishaan Gulrajani, Tianyi Zhang, Yann Dubois, Xuechen Li, Carlos Guestrin, Percy Liang, and Tatsunori~B Hashimoto. 2023.
\newblock Alpaca: A strong, replicable instruction-following model.
\newblock \emph{Stanford Center for Research on Foundation Models.}, 3(6):7.

\bibitem[{Team et~al.(2024)Team, Riviere, Pathak, Sessa, Hardin, Bhupatiraju, Hussenot, Mesnard, Shahriari, Ram{\'e} et~al.}]{team2024gemma}
Gemma Team, Morgane Riviere, Shreya Pathak, Pier~Giuseppe Sessa, Cassidy Hardin, Surya Bhupatiraju, L{\'e}onard Hussenot, Thomas Mesnard, Bobak Shahriari, Alexandre Ram{\'e}, et~al. 2024.
\newblock Gemma 2: Improving open language models at a practical size.
\newblock \emph{arXiv preprint arXiv:2408.00118}.

\bibitem[{Toshevska and Gievska(2021)}]{toshevska2021review}
Martina Toshevska and Sonja Gievska. 2021.
\newblock A review of text style transfer using deep learning.
\newblock \emph{IEEE Transactions on Artificial Intelligence}, 3(5):669--684.

\bibitem[{Tunstall et~al.(2023)Tunstall, Beeching, Lambert, Rajani, Rasul, Belkada, Huang, von Werra, Fourrier, Habib, Sarrazin, Sanseviero, Rush, and Wolf}]{tunstall2023zephyr}
Lewis Tunstall, Edward Beeching, Nathan Lambert, Nazneen Rajani, Kashif Rasul, Younes Belkada, Shengyi Huang, Leandro von Werra, Clémentine Fourrier, Nathan Habib, Nathan Sarrazin, Omar Sanseviero, Alexander~M. Rush, and Thomas Wolf. 2023.
\newblock \href {https://arxiv.org/abs/2310.16944} {Zephyr: Direct distillation of lm alignment}.
\newblock \emph{Preprint}, arXiv:2310.16944.

\bibitem[{Wei et~al.(2022)Wei, Bosma, Zhao, Guu, Yu, Lester, Du, Dai, and Le}]{wei2021finetuned}
Jason Wei, Maarten Bosma, Vincent~Y. Zhao, Kelvin Guu, Adams~Wei Yu, Brian Lester, Nan Du, Andrew~M. Dai, and Quoc~V. Le. 2022.
\newblock \href {https://openreview.net/forum?id=gEZrGCozdqR} {Finetuned language models are zero-shot learners}.
\newblock In \emph{Proceedings of ICLR}.

\bibitem[{Wieting et~al.(2017)Wieting, Mallinson, and Gimpel}]{wieting2017learning}
John Wieting, Jonathan Mallinson, and Kevin Gimpel. 2017.
\newblock Learning paraphrastic sentence embeddings from back-translated bitext.
\newblock \emph{arXiv preprint arXiv:1706.01847}.

\bibitem[{Wolf et~al.(2019)Wolf, Debut, Sanh, Chaumond, Delangue, Moi, Cistac, Rault, Louf, Funtowicz et~al.}]{wolf2019huggingface}
Thomas Wolf, Lysandre Debut, Victor Sanh, Julien Chaumond, Clement Delangue, Anthony Moi, Pierric Cistac, Tim Rault, R{\'e}mi Louf, Morgan Funtowicz, et~al. 2019.
\newblock Huggingface’s transformers: State-of-the-art natural language processing. arxiv.
\newblock \emph{arXiv preprint arXiv:1910.03771}.

\bibitem[{Xu et~al.(2023)Xu, Sun, Zheng, Geng, Zhao, Feng, Tao, and Jiang}]{xu2023wizardlm}
Can Xu, Qingfeng Sun, Kai Zheng, Xiubo Geng, Pu~Zhao, Jiazhan Feng, Chongyang Tao, and Daxin Jiang. 2023.
\newblock Wizardlm: Empowering large language models to follow complex instructions.
\newblock \emph{arXiv preprint arXiv:2304.12244}.

\bibitem[{Yang et~al.(2024)Yang, Yang, Hui, Zheng, Yu, Zhou, Li, Li, Liu, Huang et~al.}]{yang2024qwen2}
An~Yang, Baosong Yang, Binyuan Hui, Bo~Zheng, Bowen Yu, Chang Zhou, Chengpeng Li, Chengyuan Li, Dayiheng Liu, Fei Huang, et~al. 2024.
\newblock Qwen2 technical report.
\newblock \emph{arXiv preprint arXiv:2407.10671}.

\bibitem[{Yuan et~al.(2024)Yuan, Pang, Cho, Li, Sukhbaatar, Xu, and Weston}]{yuanself2024}
Weizhe Yuan, Richard~Yuanzhe Pang, Kyunghyun Cho, Xian Li, Sainbayar Sukhbaatar, Jing Xu, and Jason~E Weston. 2024.
\newblock Self-rewarding language models.
\newblock In \emph{ICML}.

\bibitem[{Zellers et~al.(2019)Zellers, Holtzman, Bisk, Farhadi, and Choi}]{zellers2019hellaswag}
Rowan Zellers, Ari Holtzman, Yonatan Bisk, Ali Farhadi, and Yejin Choi. 2019.
\newblock \href {https://doi.org/10.18653/v1/P19-1472} {{H}ella{S}wag: Can a machine really finish your sentence?}
\newblock In \emph{Proceedings of ACL}, pages 4791--4800.

\bibitem[{Zhang et~al.(2023)Zhang, Dong, Li, Zhang, Sun, Wang, Li, Hu, Zhang, Wu et~al.}]{zhang2023instruction}
Shengyu Zhang, Linfeng Dong, Xiaoya Li, Sen Zhang, Xiaofei Sun, Shuhe Wang, Jiwei Li, Runyi Hu, Tianwei Zhang, Fei Wu, et~al. 2023.
\newblock Instruction tuning for large language models: A survey.
\newblock \emph{arXiv preprint arXiv:2308.10792}.

\bibitem[{Zhao et~al.(2023)Zhao, Zhou, Li, Tang, Wang, Hou, Min, Zhang, Zhang, Dong et~al.}]{zhao2023survey}
Wayne~Xin Zhao, Kun Zhou, Junyi Li, Tianyi Tang, Xiaolei Wang, Yupeng Hou, Yingqian Min, Beichen Zhang, Junjie Zhang, Zican Dong, et~al. 2023.
\newblock A survey of large language models.
\newblock \emph{arXiv preprint arXiv:2303.18223}.

\bibitem[{Zhou et~al.(2023)Zhou, Lu, Mishra, Brahma, Basu, Luan, Zhou, and Hou}]{zhou2023instruction}
Jeffrey Zhou, Tianjian Lu, Swaroop Mishra, Siddhartha Brahma, Sujoy Basu, Yi~Luan, Denny Zhou, and Le~Hou. 2023.
\newblock Instruction-following evaluation for large language models.
\newblock \emph{arXiv preprint arXiv:2311.07911}.

\bibitem[{Zhou et~al.(2024)Zhou, Liu, Shao, Yue, Yang, Ouyang, and Qiao}]{zhou2024beyond}
Zhanhui Zhou, Jie Liu, Jing Shao, Xiangyu Yue, Chao Yang, Wanli Ouyang, and Yu~Qiao. 2024.
\newblock Beyond one-preference-fits-all alignment: Multi-objective direct preference optimization.
\newblock In \emph{Findings of ACL}, pages 10586--10613.

\end{thebibliography}

\newpage
\clearpage
\appendix
\section*{Appendices}
\section{Data Collection}
\label{sec:app_data_collection}
In this section, we provide a detailed explanation of our data construction process, divided into three parts: the details of constraint construction (\cref{sec:app_constraints}), the data diversity of \ourdata (\cref{sec:app_diversity}) and the data distribution of \ourdata (\cref{sec:app_distrubution}).

\subsection{Details of Constraints Construction}
\label{sec:app_constraints}


Table~\ref{tab:constraint}  presents the definitions of each constraint in our constraint set. 
It is important to note that for the ``Situation'', clarifying the subject or object, or defining the circumstances under which the instruction applies, we observed that generating this constraint independently often results in this additional constraint being too similar to the original instruction.  Therefore, we integrate it directly with the original interaction to develop a refined instruction. If selected during the combination process, instead of being added to the instruction like other constraints, it replaces the original instruction.


Among the constraints calculated using Python scripts, two categories are particularly unique: (1) Number-related categories: such as Length and Words Per Sentence, where we used NLTK~\citep{loper2002nltk} for calculation. (2) Keyword: We applied the lightweight, unsupervised keyword extraction method Yake~\citep{campos2020yake} to extract the top $3$ most significant keywords from the output text.
Table~\ref{table:case} provides an example generated after the constraint back-translation process.


\subsection{Dataset Diversity}
\label{sec:app_diversity}
We adopted $4$ widely used post-training datasets for constructing our CRAB dataset: Alpaca GPT-4, Open Assistant, Evol-Instruct, and Orca Chat. These datasets are of high quality and contain diverse data instances for generating rich constraints. 
The ``Rate'' column in Table~\ref{tab:constraint} shows the distribution of constraint types in the CRAB dataset.
Moreover, for constraints generated by the LLM, since we provided $13$ types of constraints as examples in the prompt, we analyzed the keywords of each generated constraint and extracted the top $10$ keywords for each major category, presenting them as subcategories in Figure~\ref{fig:constraint-type}.
The ``situation'' category was not included in the analysis because it is task-specific, making clustering difficult.

\subsection{Dataset Distrubution}
\label{sec:app_distrubution}
\begin{figure}[t]
    \centering
    \includegraphics[width=1.0\linewidth]{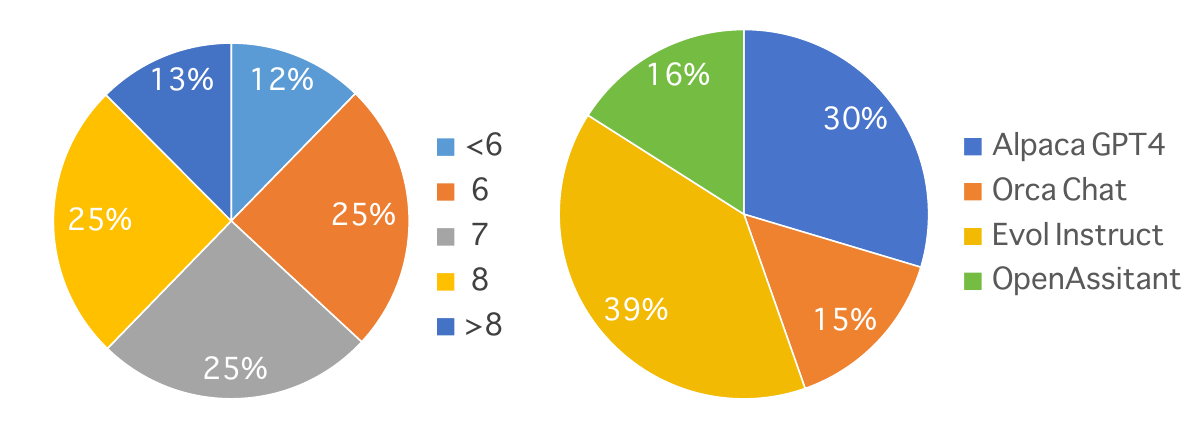} 
    \caption{ Proportion (\%) of data in the \ourdata by the number of constraints and the source dataset. 
    }
    \label{fig:distrubution}
\end{figure}

Figure~\ref{fig:distrubution} shows the distribution of $13,500$ instances in the \ourdata. The left chart categorizes data by the number of constraints after combination, while the right chart categorizes data by the source dataset. To enhance data diversity during the combination stage, we randomly introduced $25\%$ of data with a constraint count outside the $6$–$8$ range, with the maximum number of constraints being $14$.

\section{Model Training}
\label{sec:app_train}

For model training, we utilize the repository `The Alignment Handbook'~\citep{tunstall2023zephyr} to train Mistral 7B based models, and use OpenInstruct~\citep{ivison2023camels} to train LLaMA 3 8B based models.  
The implementation of the ratio \( \alpha \) between reverse training and forward training is achieved by segregating the dataset into two parts, since the model may memorize the data during the forward process.

All experiments in the paper are done using $8$ NVIDIA A$100$ $80$GB GPUs. 
We adopt DeepSpeed ZeRO stage $2$ for SFT and DPO training. The training of Mistral took approximately $48$ GPU hours in total, while the training of LLaMA3 took around $72$ GPU hours.

In the SFT stage, we set the learning rate to $5\times10^{-6}$, with a per-device batch size of $4$ and $8$ gradient accumulation steps. The warm-up ratio is set to $0.1$. The Mistral-7B experiments are trained for $4$ epochs with a maximum sequence length of $2048$, while the LLaMA 3 8B experiments are trained for $3$ epochs with a maximum sequence length of $4096$. 
For DPO training, the learning rate is set to $5\times10^{-7}$, with a per-device batch size of $4$, $2$ gradient accumulation steps, and a maximum sequence length of $2048$. The beta value for Mistral 7B experiments is set to $0.01$, trained for $1$ epoch with a cosine learning rate schedule, while for LLaMA 3 8B, the beta is $0.1$, trained for $3$ epochs with a linear learning rate scheduler following the setting in~\citet{ivison2023camels}.


\begin{table*}[!htbp]
	\centering
    \small{
	\begin{tabularx}{\textwidth}
                                    {>{\RaggedRight\arraybackslash}p{3cm} 
                                    >{\RaggedRight\arraybackslash}X 
                                    >{\centering\arraybackslash}p{1cm} 
                                    >{\centering\arraybackslash}p{1cm}} 
		\toprule
Constraint Category & Description & Generator & Rate \\ \midrule
Situation & Adding conditions, clarifying the subject or object, or defining the circumstances under which the instruction applies. & LLM & 36.9 \\ \midrule
Writing Style & Specify the style requirements for the response to align with the intended message and audience.  & LLM & 81.3 \\ \midrule
Semantic Elements & Clearly articulate the main theme, focus, meaning, or underlying concept of the response.  & LLM & 99.5 \\ \midrule
Morphological & Outline specific prohibitions, such as avoiding certain words or phrases and refraining from specific formatting styles. & LLM & 99.7 \\ \midrule
Multi-lingual & Specify the language(s). & LLM & 94.8 \\ \midrule
Literary Devices & Identify any particular literary devices to be employed. & LLM & 91.7 \\ \midrule
Grammatical Structure & Specify the grammatical structure. & LLM & 99.1 \\ \midrule
Hierarchical Instructions & Establish a response hierarchy, defining the prioritization and structuring of tasks within the output. & LLM & 83.1 \\ \midrule
Output Format & Depending on the required format of the output—such as Python, tables, JSON, HTML, LaTeX—impose relevant format constraints. & LLM & 15.2 \\ \midrule
Paragraphs Constraints & Clearly specify the required number of paragraphs or sections in the text. Additionally, indicate any specific spacing or separators needed—such as blank lines, horizontal rules, or special symbols to enhance readability and visual appeal. & LLM & 71.3 \\ \midrule
Specific Sentence & Specify a particular phrase to be included either at the beginning or end of the text, clearly indicating its exact placement. & LLM & 70.1 \\ \midrule
Header Format & Specify the formatting style for titles or keywords within the Output, such as using bold, italics, or CAPITAL LETTERS. & LLM & 9.5 \\ \midrule
Item Listing Details & Clearly specify the formatting for individual entries within the text. Direct the use of specific symbols for listing—such as bullet points (•), numbers (1., 2., 3., etc.), or hyphens (-). & LLM & 67.7 \\ \midrule
Length Constraint & Determine the word count of the output text to establish length constraints. & Python & 47.6 \\ \midrule
Word Constraint & Determine the number of words in each sentence to set word constraints. & Python & 15.1 \\ \midrule
Sentence Constraint & Determine the number of sentences in each paragraph to establish sentence constraints.  & Python & 20.6 \\ \midrule
Character Constraint & Determine the number of characters in each word. & Python & 21.4 \\ \midrule
Keyword Constraint & Determine the keywords in the output text to make the constraints more detailed. & Python & 45.1 \\ \midrule
Punctuation Limitation & Specify which punctuation marks cannot be used in the output text. & Python & 21.6 \\
\bottomrule
	\end{tabularx}%
	\caption{Constraint types defined during the back-translation process. The Rate (\%) indicates the proportion of instances in the entire dataset that generated constraints of this category.}
    \label{tab:constraint}%
    }
\end{table*}%

\section{Details on the impact of constraints on output quality}
\label{sec:app_pilot}

To explore the impact of constraints on output quality, we sampled $100$ instruction pairs from FollowBench and IFEval, with each pair consisting of a instruction without constraints and its corresponding multi-constraint version (with over $3$ constraints). Since IFEval does not provide instruction without constraints, we randomly selected $50$ instances and manually removed the constraints. For FollowBench, we selected level $0$ instructions along with their corresponding level $5$ counterparts. To ensure a fair comparison, we only retained instruction pairs where the core meaning of the instruction pairs remained consistent, such that the output generated from the complex instructions would still satisfy the simple versions.

We evaluate the quality of model output along the following four dimensions.

\begin{itemize}
    \item {\bf Engagingness: } 
    Evaluate how captivating and interesting the text is, based on its ability to hold attention and evoke interest.
    
    Components: Interest (ability to sustain attention), Appeal (suitability for the audience), and Emotional/Intellectual Impact.
    
    \item {\bf Understandability: } 
    Evaluate the clarity and ease with which the text can be understood by the target audience.
    
    Components: Simplicity (absence of unnecessary complexity), Accessibility (use of language suitable for the audience), and Clarity.

    \item {\bf Fluency: } 
    Evaluate the smoothness of the writing, focusing on grammar, sentence structure, and the natural flow of language.

    Components: Grammar (correct use of language rules), Sentence Structure (variety and complexity), and Naturalness (how easily the text flows).

    \item {\bf Coherence: }
    Evaluate the logical flow and consistency of ideas, ensuring the text’s structure is logical and ideas are connected.

    Components: Logical Flow (clear progression of ideas), Transitions (smooth movement between topics or sentences), and Consistency (absence of contradictions or disjointed thoughts).

\end{itemize}

\section{More Results}
In this section, we present additional experimental results, divided into five parts: full results on Followbench (\cref{sec:app_sharegpt}), a fairer comparison where Conifer is replaced with the same backbone as ours (\cref{sec:app_backbone}), our experimental results on LLaMA3.2-3B (\cref{sec:app_llama3.2}), the performance of general tasks (\cref{sec:general_benchmarks}), a discussion about the effect of the effect of syntactic constraints (\cref{sec:app_syn}) and a discussion about reverse training (\cref{sec:discuss_rt}).

\subsection{SSR results on Followbench}
\label{sec:app_sharegpt}

We report FollowBench results under the Hard Satisfaction Rate (HSR) metric in Table~\ref{tab:main_exp}.
Table~\ref{tab:sharegpt} presents results on FollowBench under Soft Satisfaction Rate (SSR) metric.
We also conducted a comparison with the ShareGPT version, which is trained exclusively on the ShareGPT dataset.

\begin{table*}[ht]
\centering
\small{
\resizebox{\linewidth}{!}{
\begin{tabular}{llrrrrrrrrrrrr}
\toprule
\multirow{2}{*}{Model} & \multirow{2}{*}{Backbone} & \multicolumn{5}{c}{IFEval} & \multicolumn{6}{c}{FollowBench (SSR)} & \multirow{2}{*}{AVG}  \\ \cmidrule{3-13} 
 &  & [S]P & [S]I & [L]P & [L]I & AVG & L1 & L2 & L3 & L4 & L5 & AVG &   \\ \midrule
Llama3-ShareGPT* & Llama3 & $23.7$ & $26.4$ & $33.8$ & $37.1$ & $30.3$ & $44.0$ & $40.0$ & $39.6$ & $33.3$ & $33.6$ & $38.1$ & $34.2$ \\
\llamacrab & Llama3 & $39.4$ & $50.2$ & $43.8$ & $54.2$ & $46.9$ & $57.5$ & $52.4$ & $51.2$ & $47.0$ & $45.6$ & $50.7$ & $48.8$ \\
\llamacrab + DPO & Llama3 & $40.3$ & $52.0$ & $47.7$ & $58.9$ & $49.7$ & $64.6$ & $55.8$ & $54.7$ & $52.4$ & $54.0$ & $56.3$ & $53.0$ \\ 
\midrule
Mistral-ShareGPT$\dagger$ & Mistral & $37.5$ & $49.3$ & $43.4$ & $54.9$ & $46.3$ & $55.7$ & $56.6$ & $53.6$ & $53.4$ & $49.7$ & $53.8$ & $50.0$ \\
\mistralcrab & Mistral & $47.9$ & $57.3$ & $51.6$ & $61.2$ & $54.5$ & $63.9$ & $60.6$ & $55.1$ & $50.4$ & $49.4$ & $55.9$ & $55.2$ \\ 
\mistralcrab + DPO & Mistral & $49.7$ & $61.5$ & $57.7$ & $68.5$ & $59.4$ & $66.1$ & $59.2$ & $59.8$ & $55.3$ & $51.2$ & $58.3$ & $58.8$
\\ \bottomrule
\end{tabular}
}
}
\caption{Full results (\%) on IFEval and FollowBench, where $\dagger$ and * indicate that the results are sourced from ~\citet{sun2024conifer} and ~\citet{dong2024self}, respectively.}
\label{tab:sharegpt}
\end{table*}

\begin{table*}[ht]
\centering
\small{
\resizebox{\linewidth}{!}{
\begin{tabular}{llrrrrrrrrrrrr}
\toprule
\multirow{2}{*}{Model} & \multirow{2}{*}{Backbone} & \multicolumn{5}{c}{IFEval} & \multicolumn{6}{c}{FollowBench (HSR)} & \multirow{2}{*}{AVG}  \\ \cmidrule{3-13} 
 &  & [S]P & [S]I & [L]P & [L]I & AVG & L1 & L2 & L3 & L4 & L5 & AVG &   \\ \midrule
Conifer\textsubscript{SFT}-7B$\dagger$ & Mistral & $45.8$ & $57.1$ & $50.8$ & $62.0$ & $53.9$ & $54.3$ & $49.5$ & $49.3$ & $40.8$ & $30.5$ & $44.9$ & $49.4$ \\
Conifer\textsubscript{SFT}-7B-v0.3 & Mistral & $45.8$ & $57.0$ & $49.7$ & $60.8$ & $53.3$ & $60.6$ & $52.2$ & $46.7$ & $38.8$ & $26.5$ & $45.0$ & $49.1$ \\
Conifer\textsubscript{DPO}-7B$\dagger$ & Mistral & $48.1$ & $59.1$ & $52.3$ & $63.3$ & $55.7$ & $60.3$ & $53.6$ & $48.0$ & $47.1$ & $41.0$ & $50.0$ & $52.9$ \\
Conifer\textsubscript{DPO}-7B-v0.3 & Mistral & $46.4$ & $57.2$ & $54.9$ & $64.6$ & $55.8$ & $60.1$ & $52.5$ & $46.6$ & $45.7$ & $38.6$ & $48.7$ & $52.2$ \\ \midrule
\mistralcrab + DPO & Mistral & $49.7$ & $61.5$ & $57.7$ & $68.5$ & $59.4$ & $66.1$ & $59.2$ & $59.8$ & $55.3$ & $51.2$ & $58.3$ & $58.8$
\\ \midrule
Llama3.2 3B & Llama3.2 & $15.0$ & $26.3$ & $15.5$ & $26.7$ & $20.9$ & $12.7$ & $14.7$ & $14.9$ & $18.2$ & $11.8$ & $14.5$ & $17.7$ \\
Llama3.2\textsubscript{\textsc{Crab}}\xspace & Llama3.2 & $34.9$ & $44.6$ & $38.1$ & $48.1$ & $41.4$ & $51.7$ & $36.4$ & $29.1$ & $19.7$ & $14.3$ & $30.2$ & $35.8$ \\
\bottomrule
\end{tabular}
}
}
\caption{Experimental results (\%) of the original Conifer paper, our reproduced results on Mistral 7B v0.3 (the backbone used in \mistralcrab) and the results of Llama3.2 as the backbone for IFEval and FollowBench. Here, $\dagger$ indicates that the results are from \citet{sun2024conifer}.}
\label{tab:conifer}
\end{table*}

\begin{table*}[!htb]
\centering
\small{
\resizebox{\linewidth}{!}{
\begin{tabular}{lrrrrrrrr}
\toprule
Model & \multicolumn{1}{c}{MMLU} & \multicolumn{1}{c}{GSM8k} & \multicolumn{1}{c}{MathQA} & \multicolumn{1}{c}{ARC challenge} & \multicolumn{1}{c}{ARC easy} & \multicolumn{1}{c}{Hellaswag} & \multicolumn{1}{c}{SIQA} & \multicolumn{1}{c}{AVG} \\
\midrule
Mistral-7B-v0.3 & $62.3$ & $37.2$ & $35.4$ & $48.8$ & $79.5$ & $60.9$ & $46.0$ & $52.9$ \\
\mistralcrab & $59.8$ & $46.6$ & $34.4$ & $47.4$ & $77.5$ & $62.5$ & $47.7$ & $53.7$ \\
\mistralcrab + DPO & $61.0$ & $34.0$ & $35.7$ & $53.2$ & $79.8$ & $68.1$ & $51.0$ & $54.6$ \\
\bottomrule
\end{tabular}
}
}
\caption{Experimental results (\%) on general benchmarks.}
\label{tab:general}
\end{table*}

\begin{table*}[!htb]
\centering
\small{
\resizebox{\linewidth}{!}{
\begin{tabular}{lrrrrrrrrrrrr}
\toprule
\multirow{2}{*}{Model} & \multicolumn{5}{c}{IFEval} & \multicolumn{6}{c}{FollowBench (HSR)} & \multirow{2}{*}{AVG}  \\ \cmidrule{2-12} 
 & [S]P & [S]I & [L]P & [L]I & AVG & L1 & L2 & L3 & L4 & L5 & AVG &   \\ \midrule
\mistralcrab (\texttt{w/o syntactic constraints}) & $44.2$ & $54.2$ & $47.5$ & $57.9$ & $50.9$ & $55.6$ & $50.1$ & $33.9$ & $25.1$ & $17.8$ & $36.5$ & $43.7$ \\
\midrule
\mistralcrab (\texttt{mixing throughout the process}) & $46.0$ & $56.4$ & $49.9$ & $60.1$ & $53.1$ & $60.6$ & $49.1$ & $36.5$ & $29.4$ & $15.6$ & $38.2$ & $45.7$ \\
\mistralcrab (\texttt{keeping them separate}) & $47.9$ & $57.3$ & $51.6$ & $61.2$ & $54.5$ & $63.9$ & $54.4$ & $40.1$ & $30.4$ & $27.9$ & $43.3$ & $48.9$ \\
\bottomrule
\end{tabular}
}
}
\caption{Experimental results (\%) of different mixing strategies and the effect of syntactic constraints.}
\label{tab:discuss_rt}
\end{table*}

\subsection{Different Backbone Comparison}
\label{sec:app_backbone}

\looseness=-1
Since ~\citet{sun2024conifer} did not specify the model version of Conifer, we reproduced Conifer on Mistral-7B-v0.3, which is the backbone used in \mistralcrab, and the results are presented in Table~\ref{tab:conifer}. All conclusions remain consistent with those stated in the main text.

\subsection{More Backbone Models}
\label{sec:app_llama3.2}

\looseness=-1
To verify the scalability and applicability of our approach, we conducted experiments using LLaMA3.2-3B~\citep{dubey2024llama} as the backbone, with all training hyperparameters consistent with those of LLaMA3-8B. As shown in Table~\ref{tab:conifer}, CRAB significantly improves the model's performance on complex instruction-following tasks. 

\subsection{Results on General Benchmarks}
\label{sec:general_benchmarks}

\looseness=-1
We evaluated our model on a general benchmark. 
We select several widely-used benchmarks for assessing general capabilities: MMLU~\citep{Hendrycks2020MeasuringMM}, GSM8k~\citep{cobbe2021training}, MathQA~\citep{amini2019mathqa} and Commonsense Reasoning (including HellaSwag~\citep{zellers2019hellaswag}, SIQA~\citep{sap2019social}, ARC easy and challenge~\citep{clark2018think}).
As shown in Table~\ref{tab:general}, our models improve the average performance of the original Mistral model. Especially on commonsense reasoning tasks (ARC challenge, ARC easy, HellaSwag, SIQA), \mistralcrab + DPO achieved an average improvement of $4.25\%$, showing that enhanced instruction-following boosts reasoning performance.

\subsection{Discussion about the effect of syntactic constraints}
\label{sec:app_syn}
Syntactic constraints, such as dependency relations, are common in benchmarks and user queries. Including them in our dataset improves the model's instruction-following ability, as evidenced by the performance drop when these constraints are removed, as shown in Table~\ref{tab:discuss_rt}.

\subsection{Discussion about Reverse Training}
\label{sec:discuss_rt}

Intuitively, reverse training enhances the model's understanding of constraints by requiring it to generate them, as this process inherently demands a deeper comprehension. 
We chose to keep forward and reverse data separately following~\citet{berglund2023reversal}, which separates data in their ``both'' setting. We also experimented with mixing both types of training throughout the process, and the results shown in Table~\ref{tab:discuss_rt} indicate that keeping them separate leads to better model performance. A possible reason may be that mixing the forward and reverse training may lead to overfitting easily. 



\begin{figure*}[t]
    \includegraphics[width=1.0\linewidth]{latex/figs/constraint-type.pdf} 
    \caption{The Top-$10$ subcategories for $12$ major constraint types (excluding "situation") generated by the LLM after constraint back-translation, as the "situation" type is closely tied to the input and lacks clustering tendencies.
    }
    \label{fig:constraint-type}
\end{figure*}

\begin{table*}[!htbp]
    \scalebox{0.9}{
    \begin{tabularx}{\textwidth}{X}
        \toprule
        \colorbox {pink}{\textbf{Instruction}} \\
        What is an easy J Perm PLL algorithm that wouldn’t be too complicated for a beginner at speed solving?\\
    
       \colorbox {pink}{\textbf{Output}} \\
       One easy J Perm PLL algorithm that is not too complicated for a beginner at speed solving is the following: \\
\\
R' U R' U' J R U' R' U' J' U2 R U \\
\\
This algorithm only has 10 moves and does not require any complicated finger tricks. Here is a breakdown of the algorithm:\\
\\
1. R' U R' U' - This is a common setup move for J Perm PLL algorithms. It sets up the J shape in the corner of the cube.\\
2. J R U' R' U' - This is the J Perm algorithm itself. It swaps the two adjacent edges and the two adjacent corners in the J shape.\\
3. J' U2 R U - This is the inverse of the setup move and completes the PLL algorithm.\\
\\
Remember to practice this algorithm slowly at first, and then gradually increase your speed. With enough practice, you should be able to execute this algorithm quickly and efficiently.\\
            \midrule
        \colorbox {yellow}{\textbf{Additional Constraints}} \\
    \quad \textbf{Situation:} What is a simple, $10$-move or fewer J Perm PLL algorithm for speed solving that a beginner can learn and execute efficiently, without requiring advanced finger tricks or complex sequences? \\
        \quad \textbf{Length Constraint:} Submit an response that contains at least $160$ words.\\
                \quad \textbf{Word Per Sentence:} Restrict each sentence to $25$ words maximum. \\
                \quad \textbf{Characters Per Word:} Each word in the response must consist of at least $15$ characters. \\
                \quad \textbf{Keyword Constraint:} In your reply, include the terms \"complicated finger tricks\", \"Perm PLL\", \"Perm PLL algorithm\" to emphasize the critical components and provide clarity. \\
                \quad \textbf{Punctuation Limitation:} Avoid using question marks in your response. \\
                \quad \textbf{Desired Writing Style: } The response should be written in a clear and concise manner, with a tone that is informative and helpful, suitable for a beginner in speed solving. \\
                \quad \textbf{Semantic Elements:} The main theme of the response should focus on providing a simple J Perm PLL algorithm for beginners, highlighting its ease of use and simplicity. \\
                \quad \textbf{Morphological Constraints:} Avoid using overly technical jargon or complex mathematical notation, and refrain from using formatting styles that may confuse the reader. \\
                \quad \textbf{Multi-lingual Constraints: }The response should be written in English.\\
                \quad \textbf{Specific Literary Devices:} Use a step-by-step breakdown of the algorithm to enhance clarity and facilitate understanding. \\
                \quad \textbf{Specific Grammatical Structure:} The response should be written primarily in simple sentences, with occasional use of compound sentences to provide additional explanations. \\
                \quad \textbf{Hierarchical Instructions: }The response should prioritize explaining the algorithm, followed by a breakdown of the algorithm, and finally providing practice tips. \\
                \quad \textbf{Paragraphs Constraints: }The response should consist of three paragraphs, with a blank line separating each paragraph. \\
                \quad \textbf{Specific Sentence:} The response should start with a sentence that introduces the algorithm, and end with a sentence that encourages practice. \\
                \quad \textbf{Item Listing Details: }The breakdown of the algorithm should use numbers ($1$., $2$., $3$., etc.) to list the steps, with a hyphen and a space before each step description.
        \\
       \bottomrule
    \end{tabularx}
    }
    \caption{\label{table:case} An example from OpenAssistant of \ourdata after constraint back-translation and before combination.}
\end{table*}

\end{document}